\documentclass[10pt,twocolumn,letterpaper]{article}

\usepackage[pagenumbers]{cvpr}              
\usepackage{graphicx}
\usepackage{amsmath}
\usepackage{amssymb}
\usepackage{booktabs}
\usepackage{float}
\usepackage{cuted}
\usepackage{capt-of}
\usepackage{multirow,bigstrut}
\usepackage{bbding}
\usepackage{xcolor}
\usepackage{overpic}
\usepackage{picture}
\usepackage[linewidth=1pt]{mdframed}

\usepackage[pagebackref,breaklinks,colorlinks]{hyperref}

\usepackage[capitalize]{cleveref}
\usepackage{balance}
\usepackage{algorithm}
\usepackage{color}
\usepackage{listings}
\usepackage{array}
\usepackage{bbding}
\crefname{section}{Sec.}{Secs.}
\Crefname{section}{Section}{Sections}
\Crefname{table}{Table}{Tables}
\crefname{table}{Tab.}{Tabs.}

\def\cvprPaperID{2767} 
\def\confName{CVPR}
\def\confYear{2023}

\begin{document}


\title{\noindent Real-time 6K Image Rescaling with Rate-distortion Optimization}

\author{Chenyang Qi$^{1}$\thanks{Equal contribution.} 
\qquad Xin Yang$^{*12}$ 
\qquad Ka Leong Cheng$^{1}$
\qquad Ying-Cong Chen$^{12}$
\qquad Qifeng Chen$^{1}$\\
$^{1}$HKUST \qquad $^{2}$HKUST(GZ)
}


\maketitle

\begin{abstract}
Contemporary image rescaling aims at embedding a high-resolution (HR) image into a low-resolution (LR) thumbnail image that contains embedded information for HR image reconstruction. Unlike traditional image super-resolution, this enables high-fidelity HR image restoration faithful to the original one, given the embedded information in the LR thumbnail. However, state-of-the-art image rescaling methods do not optimize the LR image file size for efficient sharing and fall short of real-time performance for ultra-high-resolution (\eg, 6K) image reconstruction.
To address these two challenges, we propose a novel framework (HyperThumbnail) for real-time 6K rate-distortion-aware image rescaling. 
Our framework first embeds an HR image into a JPEG LR thumbnail by an encoder with our proposed quantization prediction module, which minimizes the file size of the embedding LR JPEG thumbnail while maximizing HR reconstruction quality. Then, an efficient frequency-aware decoder reconstructs a high-fidelity HR image from the LR one in real time. Extensive experiments demonstrate that our framework outperforms previous image rescaling baselines in rate-distortion performance and can perform 6K image reconstruction in real time.
\end{abstract}

\section{Introduction}

\begin{figure}[ht]
    \centering
    \includegraphics[width=1.0\linewidth, trim=50 250 329 85, clip]{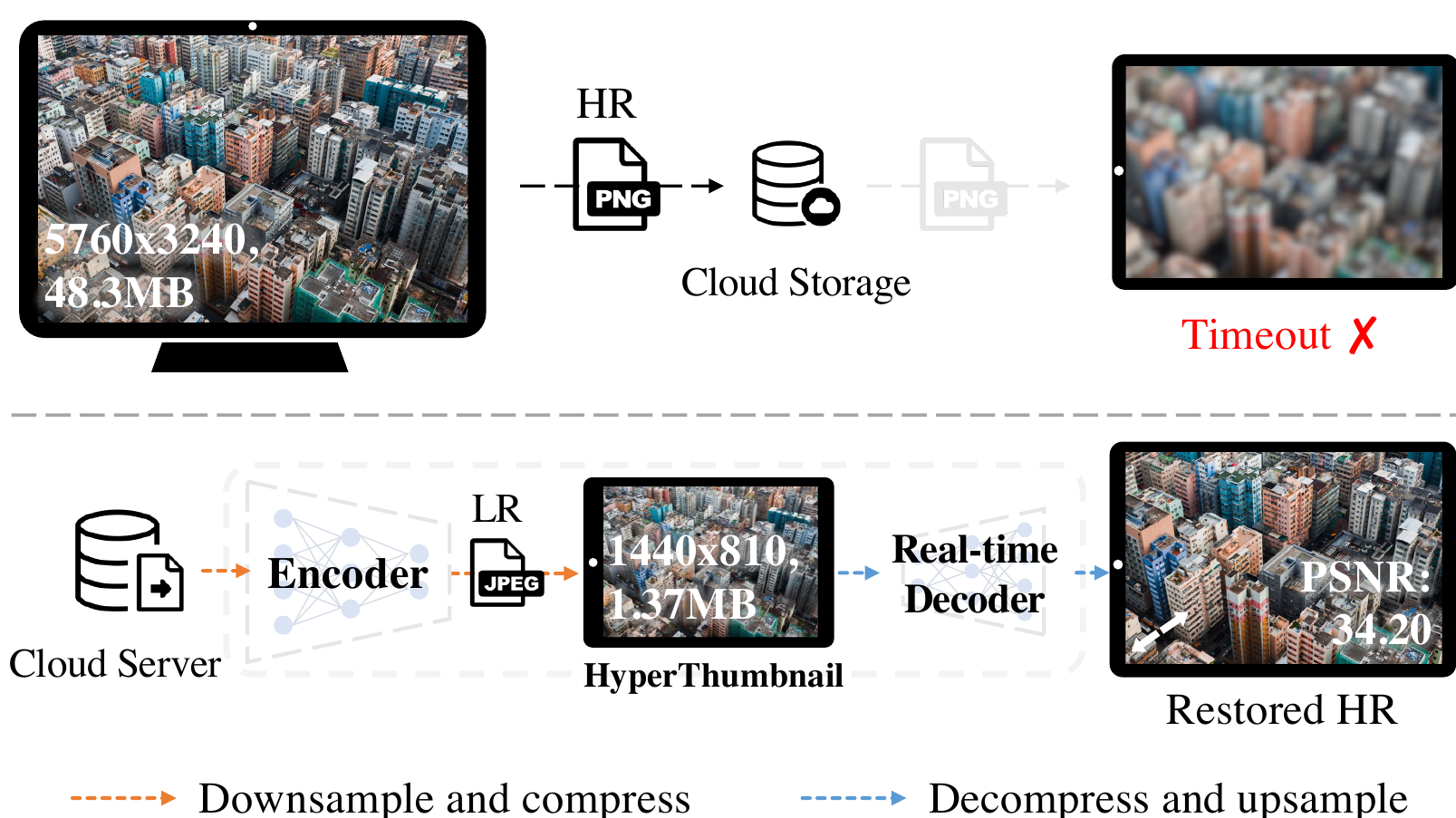}
    \caption{\textbf{The application of 6K image rescaling in the context of cloud photo storage on smartphones (\eg, iCloud).} As more high-resolution (HR) images are uploaded to cloud storage nowadays, challenges are brought to cloud service providers (CSPs) in fulfilling latency-sensitive image reading requests (\eg, zoom-in) through the internet. To facilitate faster transmission and high-quality visual content, our HyperThumbnail framework helps CSPs to encode an HR image into an LR JPEG thumbnail, which users could cache locally. When the internet is unstable or unavailable, our method can still reconstruct a high-fidelity HR image from the JPEG thumbnail in real time.}
    \label{fig:task_overview}
    \vspace{-1em}
\end{figure}

\begin{table*}[h]
\centering

\resizebox{\textwidth}{!}{
\begin{tabular}{@{}m{4cm}<{\raggedright}*{3}{m{5cm}<{\centering}}m{1cm}<{\centering} }
\toprule

Method & (a) Downsampled JPEG + super-resolution \cite{lim2017enhanced}& (b) Flow-based rescaling \cite{xiao2020invertible,liang2021hierarchical} & (c) Ours \\
\midrule
Architecture &
\includegraphics[width=0.8\linewidth]{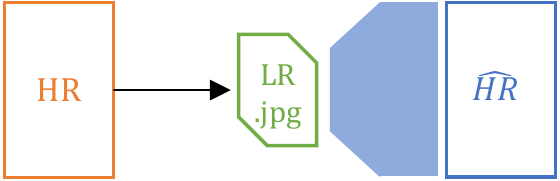} &
\includegraphics[width=0.8\linewidth]{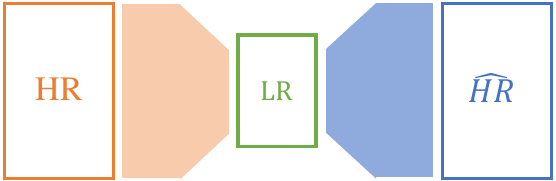} &
\includegraphics[width=0.8\linewidth]{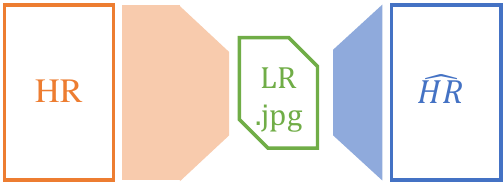}
\\

\midrule

{Reconstruction fidelity}  &
\textcolor{red}{\XSolidBrush}&
  \textcolor{teal}{\Checkmark}& 
  \textcolor{teal}{\Checkmark}\\


{Rate-distortion optimization} & 
\textcolor{red}{\XSolidBrush}&
\textcolor{red}{\XSolidBrush}& 
\textcolor{teal}{\Checkmark} \\

{{{Real-time 6K reconstruction}}} & 
--& 
\textcolor{red}{\XSolidBrush}& \textcolor{teal}{\Checkmark} \\

\bottomrule

\end{tabular}}
\caption{\textbf{The comparison of different methods related to image rescaling.} 
(a) 
Super-resolution from downsampled JPEG does not optimize rate-distortion performance and can hardly maintain high fidelity due to information lost in downsampling. (b) SOTA flow-based image rescaling methods also ignore the file size constraints and are not real-time for 6K reconstruction due to the limited speed of invertible networks. (c) Our framework optimizes rate-distortion performance while maintaining high-fidelity and real-time 6K image rescaling.
}
\vspace{-8pt}
\label{tab:comp_graph}
\end{table*}

With an increasing number of high-resolution (HR) images being produced and shared by users on the internet,
a new challenge has arisen: how can we store and transfer HR images efficiently? Storing HR images on the cloud, such as iCloud, is becoming a widely adopted solution that saves storage on a user's mobile device (\eg, smartphones) as only their low-resolution (LR) counterparts are stored on the mobile device for an instant preview. However, when a user wants to obtain the full-resolution image, the entire HR image must be downloaded on the fly from the cloud, which can result in a poor user experience when the internet connection is unstable or not available.

Real-time image rescaling can serve as a competitive solution to improving the user experience of cloud photo storage, as shown in Fig.~\ref{fig:task_overview}. 
Such a solution can first embed an HR image (on the cloud) into an LR JPEG thumbnail (on the mobile device) by an encoder, and the thumbnail provides an instant preview with little storage. When the user wants to zoom in on the thumbnail, the HR image with fine details can be reconstructed locally in real time.
In addition, image rescaling has other applications in image sharing, as it can ``bypass'' the resolution limitation of some platforms (\eg, WhatsApp) to reconstruct a high-quality HR image from an LR one \cite{xing2022scalearbitrary}. 
While modern smartphones and cameras can capture ultra-high-resolution images in 4K (iPhone 13) or even 6K (Blackmagic camera), we are interested in designing a real-time image rescaling framework for ultra-high-resolution images (\eg, 4K or 6K), which minimizes LR file size while maximizing HR and LR image quality.

However, existing image rescaling methods have their own flaws in practice, as shown in Table \ref{tab:comp_graph} where we compare different image rescaling methods in terms of their properties. One potential solution is to upsample the downsampled thumbnail with super-resolution (SR) methods \cite{dong2016image,lim2017enhanced,zhang2018residual,zhang2018image,dai2019second,liang2021swinir} (Table~\ref{tab:comp_graph}(a)). However, such a framework applies a simple downsampling strategy (\eg, Bilinear, Bicubic) to the HR image so that high-frequency details are basically lost in the LR thumbnail. Also, SR methods only focus on HR reconstruction, which leads to a sub-optimal image rescaling performance.
Instead, dedicated image rescaling approaches aim to embed information into a visually pleasing LR image and then reconstruct the HR image with an upsampling module.
Recently, state-of-the-art image rescaling works utilize normalizing flow~\cite{xiao2020invertible,liang2021hierarchical,huang2021video,xing2022scalearbitrary} show impressive image embedding and reconstruction capability that outperforms SR approaches, in terms of the reconstructed HR image quality.
However, there are still some great challenges to apply these flow-based rescaling frameworks in real-world applications, as shown in Table~\ref{tab:comp_graph}(b). First, the file size of the LR thumbnail is not optimized.
Second, the reconstruction stage of these image rescaling methods is computationally expensive due to their invertible network architecture with extensive use of dense blocks~\cite{huang2017densely}: IRN~\cite{xiao2020invertible} costs about a second to reconstruct a 4K image with 4x rescaling on a modern GPU, which is far from real time (Table~\ref{table:efficiency and fidelity}).



In this work, we propose the \textbf{HyperThumbnail}, a rate-distortion-aware framework for 6K real-time image rescaling, as shown in Table~\ref{tab:comp_graph}(c). 
In this framework, we embed an HR image into a low-bitrate JPEG thumbnail by an encoder and a quantization table predictor, as JPEG is a dominant image compression format today~\cite{JPEGUsageStatistics}. Then the JPEG thumbnail can be upscaled to its high-fidelity HR counterpart with our efficient decoder in real time. 
We leverage an asymmetric encoder-decoder architecture, where most computation is put in the encoder to keep the decoder lightweight. This makes it possible for our decoder to upscale a thumbnail to 6K in real time, significantly faster than previous flow-based image rescaling methods~\cite{xiao2020invertible, liang2021hierarchical}.

Meanwhile, the Rate-Distortion (RD) performance is an important and practical metric rarely studied in prior rescaling works. 
In this paper, we define the \textbf{rate} as the ratio between the thumbnail file size and the number of pixels in the HR image, also known as the bits-per-pixel (bpp). The \textbf{distortion} consists of two parts: the perceptual quality of the thumbnail (LR distortion) and the fidelity of the restored HR image (HR distortion). 
The rate-distortion performance evaluates an image rescaling framework in both storage cost and visual quality.
Without explicit RD constraints, recent works in image rescaling ~\cite{xiao2020invertible, liang2021hierarchical} do not consider RD performance in their models. While some works~\cite{wu2021embedding,xing2021invertible,jiang2018end,son2022enhanced,xiao2023invertible} leverage the rate constraint by embedding extra information in JPEG, they simply utilize a fixed differentiable JPEG module, which we argue is sub-optimal for image rescaling. 
Because such a process deteriorates the information in the embedding images without considering their local distribution. 
Moreover, the quantization process of JPEG introduces noise in the frequency domain and introduces well-known JPEG artifacts, which brings great challenges to information restoration.

To remedy these issues, our image rescaling framework is designed to jointly optimize image quality and bpp with entropy models.
Instead of using fixed quantization tables in conventional JPEG (Sec.~\ref{sec:JPEG_algo}), we propose a novel quantization prediction module (QPM) that predicts image-adaptive quantization tables, which can optimize RD performance.
We further adopt a frequency-aware decoder which alleviates JPEG artifacts in the thumbnails and improves HR reconstruction. 
Moreover, our asymmetric encoder-decoder framework can be extended to optimization-based compression.

Our contributions are summarized as follows:
\begin{itemize}
    \renewcommand{\labelitemi}{\textbullet}
    \item We propose a 6K real-time rescaling framework with an asymmetric encoder-decoder architecture, named HyperThumbnail, which embeds a high-resolution image into a JPEG thumbnail that can be viewed in popular browsers. The decoder utilizes both spatial and frequency information to reconstruct high-fidelity images in real time for 6K image upsampling.
    \item We introduce a new quantization prediction module (QPM) that improves the RD performance in the encoding stage of our framework. Furthermore, we adopt rate-distortion-aware loss functions along with QPM to optimize the RD performance.
    \item Experiments show that our framework outperforms state-of-the-art image rescaling methods with higher LR and HR image quality and faster reconstruction speed at similar file size.
\end{itemize}

\section{Related Work}

\subsection{Image Super-resolution}

Image super-resolution (SR) targets at restoring HR images from LR images. Pioneering works on SR such as SRCNN~\cite{dong2016image}, EDSR~\cite{lim2017enhanced}, and other successors \cite{zhang2018residual,zhang2018image,dai2019second,li2020lapar,li2022mulut} exploit deep neural networks to solve the challenges in image SR. 
Recently, Liang $\etal$~\cite{liang2021swinir} propose a Swin Transformer-based framework with state-of-the-art image SR performance. 
In addition, Wang $\etal$~\cite{wang2018esrgan,wang2021real} extend SRGAN~\cite{ledig2017srgan} and demonstrate the potential of producing perceptually pleasing HR images with a GAN-based generator.
Despite great recent progress, most of these methods assume a simple deterministic downsampling process (\eg, Bicubic downsampling ~\cite{mitchell1988reconstruction}), which limits the reconstruction quality of high-frequency details in image SR.


\subsection{Image Rescaling}

Different from SR, image rescaling aims to downsample the given HR image into a visually satisfying LR image with embedded information, and then reconstruct the HR image with high fidelity.
A straightforward solution is downscaling using detail-preserving methods~\cite{kopf2013content} and upscaling with heavy SR networks~\cite{lim2017enhanced,dai2019second,liang2021swinir}.
Recently, efforts have been made to apply invertible neural network (INN) to image rescaling.
INN~\cite{dinh2015nice,dinh2017density,kingma2018glow,kumar2019videoflow,grathwohl2019ffjord,behrmann2019invertible,chen2019residual,ouyang2022restorable} provides direct access to the inverse mapping of the forward function, making it a popular framework for image rescaling~\cite{xiao2020invertible,liang2021hierarchical,cheng2021iicnet}.
Xiao $\etal$~\cite{xiao2020invertible} make the first attempt to model image downscaling and upscaling using invertible transformation. Liang $\etal$~\cite{liang2021hierarchical} further formulate the high-frequency components in INNs as a conditional distribution on LR image.
However, the equivalence in the computational cost of the encoding and decoding process of INN makes it almost impossible to optimize the backward inference time independently, which limits its practicality in latency-aware scenarios. 
Besides, the file size of the embedding LR image in existing rescaling frameworks is yet to be studied. In this paper, we leverage the RD metric to evaluate the rescaling methods from a real-world perspective and compare our framework with existing image rescaling models in terms of their bpp and restored HR fidelity. 

\subsection{Image Compression}
\label{sec:related_compression}
Instead of only shrinking the image size spatially, image compression approaches optimize the RD performance by first producing a compact bitstream, then decoding HR from the bitstream. 
Recently, learning-based methods~\cite{balle2017end,rippel2017real,balle2018variational,agustsson2019generative,minnen2018joint,toderici2017full,xie2021enhanced,kim2022joint,he2022elic, lee2022dpict} have improved RD performance by a large margin with neural encoders and decoders. 
However, these methods have not been widely adopted on the internet due to issues related to software compatibility or runtime efficiency.
If users of neural compression need a thumbnail for easier manipulation and preview, they need to save another redundant file besides the bitstream, which is inconvenient and takes extra storage. Thus, traditional compression algorithms (\eg, JPEG~\cite{wallace1991jpeg}) are still popular on most social media platforms~\cite{JPEGUsageStatistics}.

In this work, we embed HR images into LR JPEG thumbnails and restore HR images with a 6K real-time decoder. 
Training neural networks with the non-differentiable JPEG algorithm is challenging. Previous works approximate JPEG either with the iterative optimization~\cite{jiang2018end} or the differentiable degradation simulator~\cite{son2022enhanced,zhao2019learning} during training, and revert to the conventional JPEG algorithm at test time.
However, these methods do not optimize the RD performance in their models.
In a concurrent work, Xiao $\etal$~\cite{xiao2023invertible} extend IRN with a fixed JPEG compression module and an extra JPEG artifact removal module. However, they have not considered the bpp and the reconstruction quality as a joint optimization problem. 
In this work, we propose a novel differentiable JPEG process with the QPM guided by a bitrate loss for RD performance optimization. Moreover, our real-time decoder reconstructs the HR image from both the spatial and frequency domains of a thumbnail and further improves the reconstruction quality. 
\section{Method}

\begin{figure*}[ht]
\centering
\begin{tabular}{@{}c@{}}
\includegraphics[width=1.0\linewidth, trim=60 160 75 90, clip]{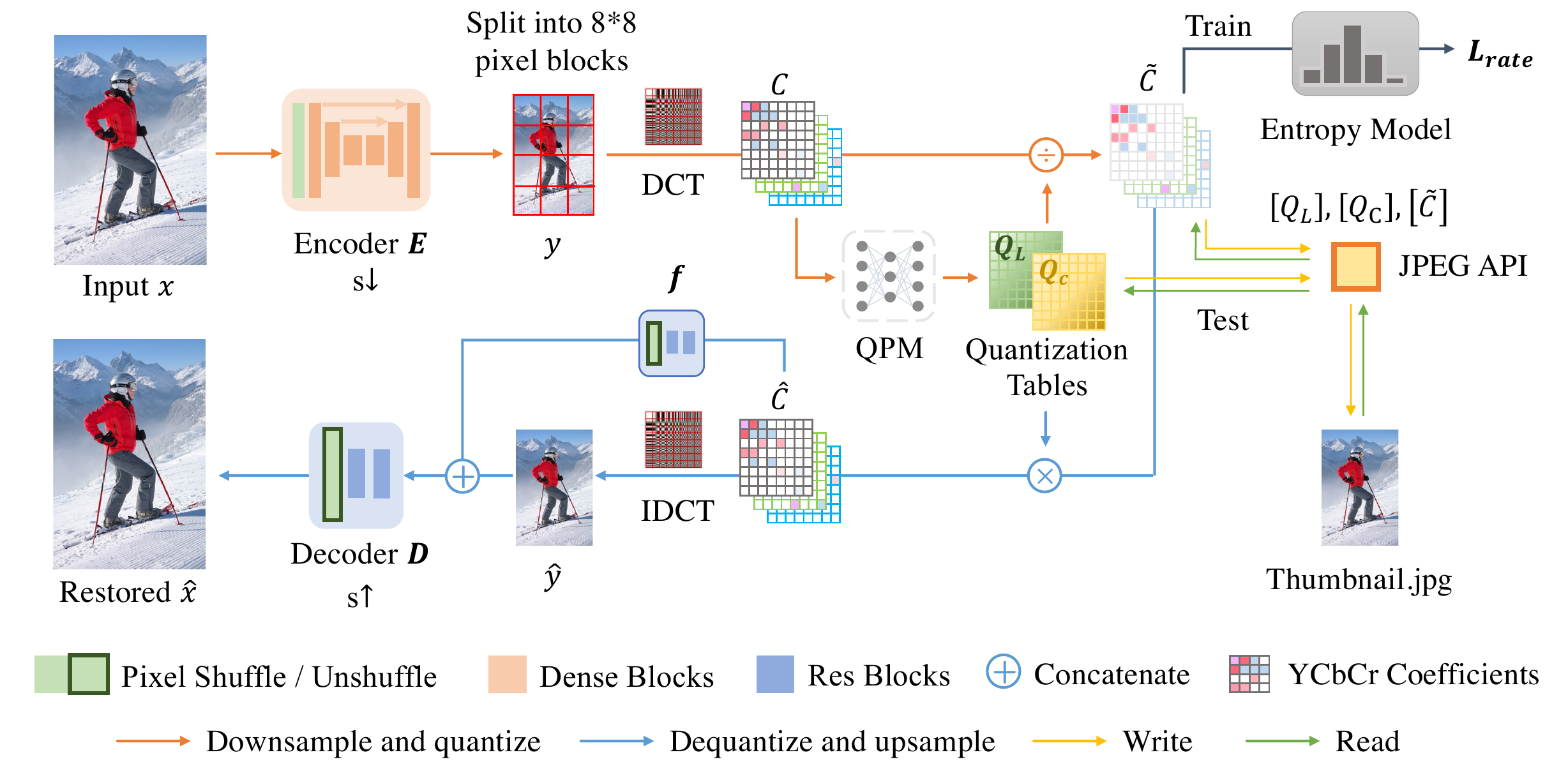}
\end{tabular}
\vspace{-0.5em}
\caption{\textbf{The overview of our approach.} Given an HR input image $x$, we first encode $x$ to its LR representation $y$ with the encoder $E$, where the scaling factor is $s$. Second, we transform $y$ to DCT coefficients $C$ and predict the quantization tables $Q_L, Q_C$ with our quantization prediction module (QPM). Third, we adopt an entropy model~\cite{balle2017end} to estimate the bitrate of the quantized coefficients $\widetilde{C}$ at training stage. After the rounding and truncation, which we denoted as $[ \cdot ]$, the $[Q_L], [Q_C]$ and $[\widetilde{C}]$ can be written and read with off-the-shelf JPEG API at the testing stage. To restore the HR, we extract features from $\widehat{C}$ with a frequency feature extractor $f$ and produce the high-fidelity image $\hat{x}$ with the decoder $D$.
}
\label{fig:framework}
\vspace{-0.5em}
\end{figure*}

\subsection{JPEG Preliminary}
\label{sec:preliminary}
\label{sec:JPEG_algo} 
As our solution involves optimizing the rate-distortion (RD) performance of the JPEG thumbnail, we briefly summarize the JPEG algorithm~\cite{wallace1991jpeg} in this section.
The JPEG algorithm compresses an image in three steps. First, given an RGB image $y \in \mathbb{R}^{3 \times h \times w}$, the algorithm converts $y$ into the luma-chroma color space (YCbCr).
Second, the converted image is divided into ${8\times8}$ pixel blocks as $y \in \mathbb{R}^{3\times N \times8\times8}$, where $N = \frac{h\times w}{8\times8}$. 
Then, $y$ is transformed to its corresponding Discrete-Cosine-Transform (DCT) coefficients $C \in \mathbb{R}^{3 \times N \times 8 \times 8}$:
\begin{equation}
    C = ({C}_{Y}, {C}_{Cb}, {C}_{Cr}) = {\rm DCT}(y_{Y}, y_{Cb}, y_{Cr}). 
\end{equation}
Third, the luma coefficient ${C}_{Y}$ and chroma coefficient $C_C = ({C}_{Cb}, {C}_{Cr})$ are quantized separately by two quantization tables $Q_L$ and $Q_C$:
\begin{equation}
\label{equ:quant_func}
\begin{aligned}
    \widetilde{C_Y}=\frac{C_Y}{[ Q_{L} ]}, \quad \widetilde{C_C} =  \frac{C_C}{[ Q_{C} ]}, \\
\end{aligned}
\end{equation}
where $[ \cdot ]$ represents the rounding and truncation function. 
Unlike the original JPEG algorithm which applies a set of image-invariant quantization tables, our work predicts the $Q_L$ and $Q_C$ with the QPM (Sec.~\ref{sec:QPM}) for different images. 
As shown in the bottom-right of Fig.~\ref{fig:framework}, the quantization tables $[Q] = ([Q_L], [Q_C])$ and quantized DCT coefficients $[\widetilde{C}] = ([\widetilde{C_Y}], [\widetilde{C_C}])$ are encoded into a JPEG file.
At the decoding phase, the JPEG decoder can extract $[Q], [\widetilde{C}]$ from the JPEG file for image reconstruction with inverted operations of above steps:
\begin{equation}
    \label{eqn:jpeg_dec}
    \widehat{C_Y} = [\widetilde{C_Y}][Q_L],\quad \widehat{C_C} =   [\widetilde{C_C}][Q_C],\\
\end{equation}
and $\hat{y} = {\rm IDCT}(\hat{C})$, where IDCT is the inverse operation of DCT, and $\hat{C}$ is the abbreviation of $[\widehat{C_Y}, \widehat{C_C}]$.

\subsection{Overview of HyperThumbnail}
Inspired by recent progress in high-fidelity efficient restoration through an efficient MLP~\cite{muller2022instant} or a small transformer~\cite{he2022masked}, we adopt an asymmetric encoder-decoder framework, which enables real-time reconstruction.
Fig.~\ref{fig:framework} illustrates the overview of our framework.
Given an HR input image 
$x \in \mathbb{R}^{3 \times H \times W}$, we first generate its LR representation
$y \in \mathbb{R}^{3 \times \frac{H}{s} \times \frac{W}{s}}$
through our encoder $E$,
where the $s$ is the rescaling factor and our encoder $E$ is an 
U-Net~\cite{ronneberger2015unet} with 
dense blocks~\cite{huang2017densely}.
To further decrease the file size, we transform $y$ to DCT coefficients $C \in \mathbb{R}^{3  \times \frac{H \times W}{64 \times s^2}\times 8 \times 8}$. Then, we quantize $C$ with image-specific quantization tables $Q \in \mathbb{R}^{2 \times 8 \times 8}$ (Sec.~\ref{sec:QPM}).
During the test time, we encode and decode $(Q, \widetilde{C})$ with off-the-shelf JPEG API, and then retrieve $(\widehat{C}, \hat{y})$ with Eqn.~\eqref{eqn:jpeg_dec}.
In the upsampling stage, we reconstruct the high-fidelity HR $\hat{x}$ with our efficient frequency-aware decoder in real time (Sec.~\ref{sec:Coefficient-wise Convolution Module}).

\subsection{Quantization Prediction Module}
\label{sec:QPM}
As JPEG~\cite{wallace1991jpeg} is a widely adopted compression algorithm, some previous methods \cite{son2022enhanced,jiang2018end,zhao2019learning,xing2021invertible} in image embedding also introduce JPEG (Sec.~\ref{sec:JPEG_algo}) to compress their encoded image. However, JPEG degrades the embedded information in the encoded image and leads to a performance drop in the decoding stage (Sec.~\ref{sec:ablation}). JPEG uses fixed quantization tables with constant values~\cite{wallace1991jpeg}, which we believe is suboptimal since different images have different frequency distribution. Hence, besides encoding HR images to LR thumbnails, we also predict the quantization tables $Q$ for each thumbnail with the quantization prediction module (QPM), which boosts the RD performance of our framework significantly. 
The QPM is implemented as two separate 8-layer multilayer perceptron (MLPs): the luma predictor ${\rm MLP}_L$ and the chroma predictor ${\rm MLP}_C$. For block $C_{k}=(C_{Y,k},C_{C,k})$, we vectorize $C_k$ into a 1D vector and produce its quantization table with QPM. Thus we have
\begin{eqnarray}
    \label{equ:ql_qc}
    \begin{aligned}
        Q_L =& \frac{\sum_k {\rm MLP}_L(C_{Y, k})}{|C_{Y, k}|}, 
        Q_C =& \frac{\sum_k {\rm MLP}_C(C_{C, k})}{|C_{C, k}|},
    \end{aligned}
\end{eqnarray}
where $| \cdot |$ denotes the block counts.
To facilitate the conventional JPEG codec, we take the average $Q$ as the quantization table for the whole image.
Following the training stage of learned compression~\cite{balle2018variational}, we adopt the additive uniform noise $\epsilon$ where each element 
follows $\mathcal{U}(-\frac{1}{2}, \frac{1}{2})$ to approximate the non-differentiable quantization noise introduced in Eqn.~\eqref{equ:quant_func}:
\begin{equation}
\label{equ:approx_quant}
\begin{aligned}
    \widetilde{C_Y}=\frac{C_Y}{Q_L + \epsilon} + \epsilon, \quad \widetilde{C_C} =  \frac{C_C}{Q_C + \epsilon} + \epsilon.
\end{aligned}
\end{equation}
In the testing stage, we switch back to the standard quantization function in Eqn.~\eqref{equ:quant_func} to fit the JPEG API. Experiments show that our QPM predicts better quantization tables $Q$ so that it improves the RD performance of our framework (Sec.~\ref{sec:ablation}).
Our experiments in the 
supplement
also show that our QPM improves the standard JPEG algorithm.

\subsection{Frequency-aware Decoder}
\label{sec:Coefficient-wise Convolution Module}
As mentioned in Sec.~\ref{sec:JPEG_algo}, the JPEG quantization is applied on the DCT domain coefficients $\widetilde{C}$ and leads to the well-known quantization noise.
Previous works
~\cite{wu2021embedding,xing2021invertible,son2022enhanced} usually decode the embedded information from the RGB domain of the JPEG image.
However, we note that quantization noise can be approximated by simple independent uniform noise~\cite{balle2017end} in the DCT domain, but it is nonlinear and more complex to model in the RGB domain after IDCT.

Inspired by recent works in JPEG artifact removal~\cite{ehrlich2020quantization,wang2016d3,zhang2018dmcnn,guo2016building} and image generation~\cite{nash2021generating}, we further propose a novel plug-in named frequency feature extractor $f$ for additional frequency domain perception. 
We first reshape the dequantized DCT coefficients into a sparse representation $\widehat{C} \in \mathbb{R}^{192 \times \frac{H}{8s}\times \frac{W}{8s}}$~\cite{nash2021generating}. 
Then, we extract features $f(\widehat{C})\in \mathbb{R}^{24 \times \frac{H}{s}\times \frac{W}{s}}$ and concatenate $f(\widehat{C})$ with the RGB image $\hat{y}$. The concatenated features are fed into the decoder to reconstruct the HR image $\hat{x}$:
\begin{eqnarray}
\hat{x} &=& D(\hat{y} \oplus f(\widehat{C})),
\end{eqnarray}
where $\oplus$ is the concatenation operator along the channel dimension.
Since $\widehat{C}$ is $\frac{1}{8}$ of $\hat{y}$ in the spatial resolution, $f$ takes negligible computational overhead. In this way, our decoder is lightweight enough to reconstruct 6K HR images in real time (Fig.~\ref{fig:trt_runtime}). $f$ improves our RD performance and is more efficient than increasing the decoder capacity. Please refer to our 
supplement
for more details of our architecture.


\subsection{Training Objectives}
\label{sec:training_obj}
\noindent \textbf{Bitrate loss}~
To optimize the RD performance of our framework, it is critical to estimate the code length (bitrate) of the quantized coefficients $\widetilde{C}$ at the training stage.
Inspired by Ball{\'e} $\etal$~\cite{balle2017end}, the rate $R$ of $\widetilde{C}$ is estimated with differentiable fully-factorized entropy models:


\vspace{0.2em}
\begin{equation}
\label{equ:entropy}
    R = \mathbb{E}_{x \sim p_x}[-\log_2 ~p_L(\widetilde{C}_Y) -\log_2 ~p_C(\widetilde{C}_{Cb}) - \log_2 ~p_C(\widetilde{C}_{Cr})],
\end{equation}
\vspace{0.2em}

where $p_L$ and $p_C$ are two fully-factorized entropy models for luma and chroma coefficient maps, respectively. 
In our work, we reshape $\widetilde{C}$ to $(3, 64, \frac{H}{8s}, \frac{W}{8s})$, and assume the value of each pixel in $\widetilde{C}$ to be independent following Ball{\'e} $\etal$~\cite{balle2017end}. For accurate rate estimation, we train $p_L$ and $p_C$ to model the 64-channel probability density functions of $\widetilde{C}_Y$ and $(\widetilde{C}_{Cb}, \widetilde{C}_{Cr})$ by minimizing an auxiliary loss $L_{aux}$~\cite{balle2018variational}. The bpp of the restored image $\hat{x}$ is calculated by 
\begin{equation}
    L_{bpp} = \frac{R}{H \times W}.
    \label{equ:l_rate}
\end{equation}
More details of auxiliary loss is provided in the supplement.

\label{sec:recon_loss}
\noindent \textbf{Reconstruction and guidance loss}~
Thanks to the fully differentiable pipeline (Fig.~\ref{fig:framework}), it is possible for us to train our encoder, QPM and decoder with similar loss terms used in previous works~\cite{xiao2020invertible,liang2021hierarchical}. 
Following IRN~\cite{xiao2020invertible}, we employ $L_1$ on the reconstructed $\hat{x}$ and a $L_2$ guidance loss on the JPEG thumbnail $\hat{y}$:
\begin{eqnarray}
\vspace{-1em}
    L_{recon} &=& \frac{||\hat{x} - x||_1}{H \times W},    \label{equ:L_recon}\\
    L_{guide} &=& \frac{||\hat{y} - y_{ref}||_{2}^2}{(H/s) \times (W/s)},
    \label{equ:L_guide}
\vspace{-1em}
\end{eqnarray}
where $y_{ref}$ is a guidance image downsampled from $x$ with bicubic interpolation \cite{mitchell1988reconstruction}.
Altogether, we train our framework by minimizing the total loss $L_{rescale}$:
\begin{eqnarray}
\vspace{-1em}
    \label{equ:l_pixel}
    L_{rescale}&=& L_{recon} + \lambda_{1}L_{guide} + \lambda_{2}L_{bpp}.
\vspace{-1em}
\end{eqnarray}
Empirically, we set $\lambda_1=0.6$. The target bpp of our framework could be adjusted by a loss scaling factor $\lambda_{2}$, which we set to $0.01$ for most experiments.
The total loss $L_{rescale}$ is adopted to optimize the parameters of the encoder, the decoder, and QPM.

\begin{table*}[t]

\centering
\resizebox{1\linewidth}{!}{
\begin{tabular}{@{}l@{\hspace{2mm}}*{2}{c@{\hspace{1mm}}}c@{\hspace{0mm}}c@{\hspace{2mm}}c@{\hspace{4mm}}c@{\hspace{4mm}}c@{\hspace{3mm}}*{5}{c@{\hspace{2mm}}}}

%
\toprule
Method & \multicolumn{3}{c}{Bitrate$\downarrow$-Distortion$\uparrow$~\cite{KodakDataset}}& \multicolumn{2}{c}{Upscaling Efficiency$\downarrow$} &
\multicolumn{6}{c}{Reconstructed HR PSNR$\uparrow$} \\
\cmidrule(l{1mm}r{1mm}){2-4} 
\cmidrule(l{1mm}r{1mm}){5-6} \cmidrule(l{1mm}r{1mm}){7-13}
Down \& Degradation \& Up & &bpp & PSNR & Time (ms) & GMacs & & Set5 &  Set14  &  BSD100  &  Urban100 &  DIV2K & FiveK-6k \\
\midrule

Bicubic \& JPEG \& Bicubic& &0.29 & 25.18 & -- & -- &  & 25.14 & 23.49 & 24.02 & 21.05 & 25.70 & 
26.90 \\
Bicubic \& JPEG \& EDSR~\cite{lim2017enhanced} & &0.29 & 26.77 & 91.0 & 1007.5 & & 28.34 & 25.73 & 25.60& 23.58 & 27.83 & 27.23 \\
Bicubic \& JPEG \& SwinIR~\cite{liang2021swinir} & & 0.29 & 26.93 & 4012.6 & 6208.7 & & 28.56 & 25.99 & 25.72 & 24.09 & 28.07 & 27.44 \\
ComCNN \& RecCNN \cite{jiang2018end} & & 0.32 & 27.02 & 469.7 & 6014.7 & & 28.29 & 25.84 & 25.78 & 23.70 & 27.99 &  27.40  \\

IRN~\cite{xiao2020invertible} \& JPEG  & & 0.31 & 28.48 & 977.8 & 4751.7 & & 30.00 & 27.23 & 26.91 & 25.72 & 29.54 
& 27.96   \\

HCFlow~\cite{liang2021hierarchical} \& JPEG & & 0.30 & 28.76 & 1025.9 & 4626.0 & & 29.98 & 27.41 & 27.05 & 26.19 & 29.71 & 
28.01\\

\midrule
Ours-full & & 0.30 & \textbf{29.67} & 247.9 & 1277.5 & &  \textbf{30.48} & \textbf{28.21} & \textbf{27.93} & \textbf{27.35} & \textbf{30.49}
& \textbf{28.51} \\
Ours & & 0.30 & 29.42  & \textbf{37.8} & 
\textbf{156.2}
& & 30.22 & 27.87 & 27.66 & 26.62 & 30.15 & 
28.15 \\


\bottomrule
\end{tabular}
}

\vspace{-1mm}
\caption{Quantitative evaluation of upscaling efficiency and reconstruction fidelity. We keep bpp around 0.3 on Kodak~\cite{KodakDataset} for different methods, and the distortion is measured by the PSNR on the reconstructed HR images. Our approach outperforms other methods with better HR reconstruction and a significantly lower runtime. We measure the running time and GMacs of all models by upscaling a 960 $\times$ 540 LR image to a 3840 $\times$ 2160 HR image. The measurements are made on an Nvidia RTX 3090 GPU with PyTorch-1.11.0 in half-precision mode for a fair comparison.}
\label{table:efficiency and fidelity}
\vspace{-1em}
\end{table*}

\begin{figure*}
\centering
\begin{tabular}{@{}c@{\hspace{2mm}}c@{\hspace{2mm}}c@{\hspace{2mm}}c@{\hspace{2mm}}c@{}}
\footnotesize{Ground Truth}& 
\footnotesize{IRN~\cite{xiao2020invertible} ~$4\times$ \& JPEG q=96}&
\footnotesize{HCFlow~\cite{liang2021hierarchical}~$4\times$ \& JPEG q=90}&
\footnotesize{Ours $4\times$}&
\footnotesize{Ours-full $4\times$}
\\
\begin{overpic}
    [width=0.19\linewidth]{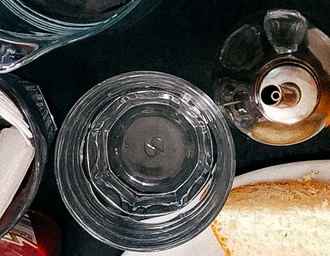}
   \put(0.14\linewidth, -1mm){ 
   \includegraphics[width=0.05\linewidth ]{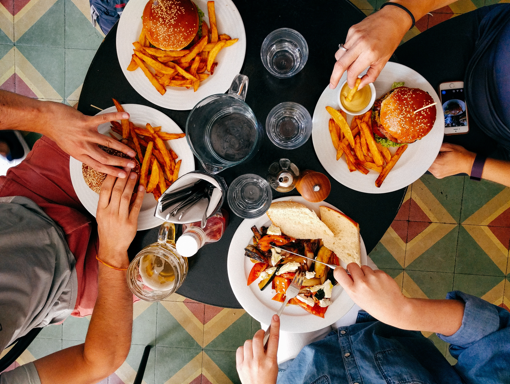}
    }
   \put(0.14\linewidth + 3.5mm, 1.5mm){
    \includegraphics[width=0.01\linewidth ]{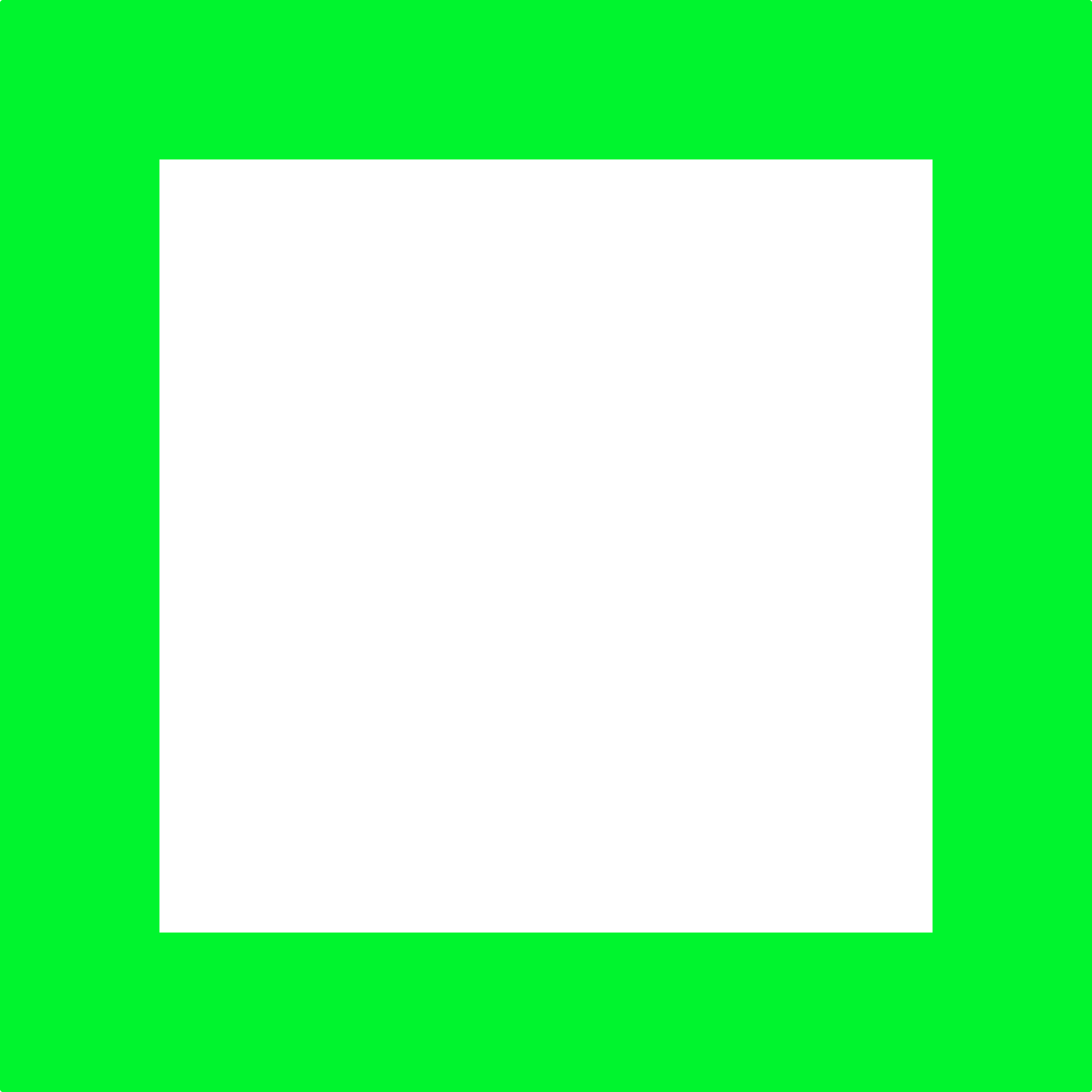}}
\end{overpic}&
\begin{overpic}
    [width=0.19\linewidth]{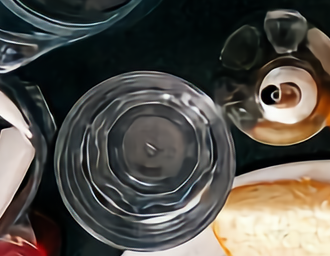}
   \put(0.14\linewidth, -1mm){
    \includegraphics[width=0.05\linewidth]{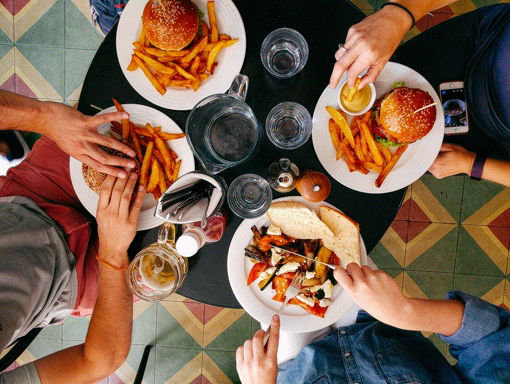}}
   \put(0.14\linewidth + 3.5mm, 1.5mm){
    \includegraphics[width=0.01\linewidth]{figures/section4_experiment/Visual_Comparison/bbox.png}}
\end{overpic}&
\begin{overpic}
    [width=0.19\linewidth]{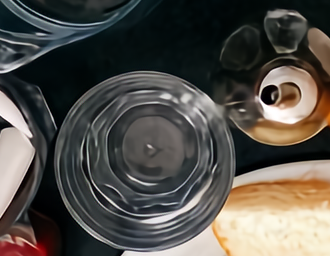}
   \put(0.14\linewidth, -1mm){
    \includegraphics[width=0.05\linewidth]{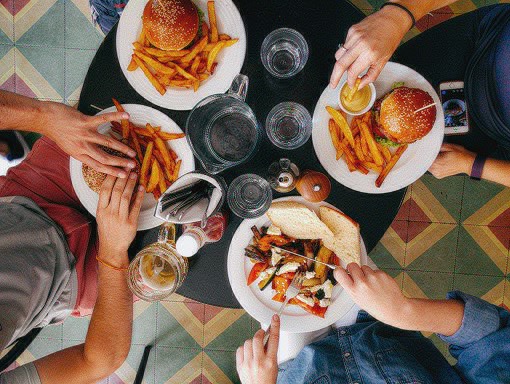}}
   \put(0.14\linewidth + 3.5mm, 1.5mm){
    \includegraphics[width=0.01\linewidth]{figures/section4_experiment/Visual_Comparison/bbox.png}}
\end{overpic}&
\begin{overpic}
    [width=0.19\linewidth]{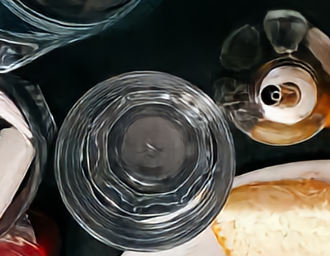}
   \put(0.14\linewidth, -1mm){
    \includegraphics[width=0.05\linewidth]{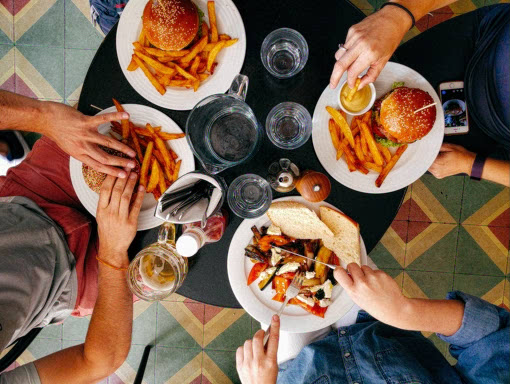}}
   \put(0.14\linewidth + 3.5mm, 1.5mm){
    \includegraphics[width=0.01\linewidth]{figures/section4_experiment/Visual_Comparison/bbox.png}}
\end{overpic}&
\begin{overpic}
    [width=0.19\linewidth]{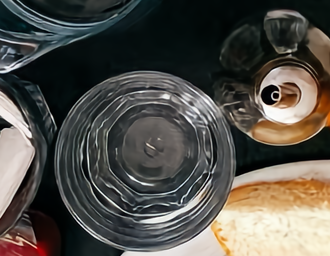}
   \put(0.14\linewidth, -1mm){
    \includegraphics[width=0.05\linewidth]{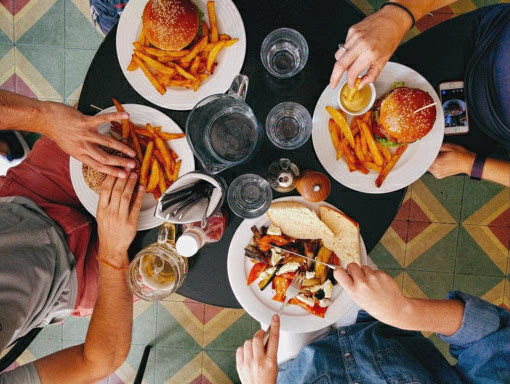}}
   \put(0.14\linewidth + 3.5mm, 1.5mm){
    \includegraphics[width=0.01\linewidth]{figures/section4_experiment/Visual_Comparison/bbox.png}}
\end{overpic}
\\
\vspace{1em}
\footnotesize{bpp$\downarrow$~/~PSNR(RGB)$\uparrow$}& 
\footnotesize{0.324~/~22.82}&
\footnotesize{0.296~/~22.47}&
\footnotesize{\textbf{0.240~/~24.32}}&
\footnotesize{\textbf{0.238~/~24.78}}
\vspace{-0.5em}
\\
\begin{overpic}
    [width=0.19\linewidth]{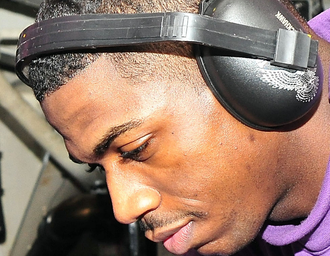}
   \put(0.14\linewidth, -1mm){
    \includegraphics[width=0.05\linewidth]{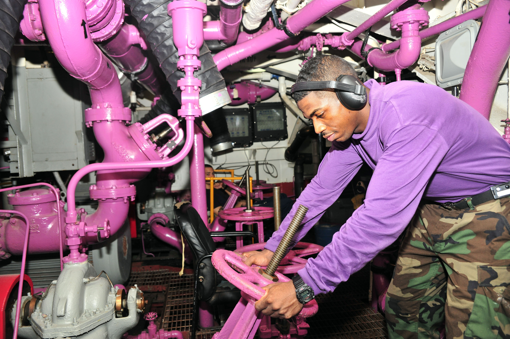}}
   \put(0.14\linewidth + 4.9mm, 2.3mm){
    \includegraphics[width=0.007\linewidth]{figures/section4_experiment/Visual_Comparison/bbox.png}}
\end{overpic}&
\begin{overpic}
    [width=0.19\linewidth]{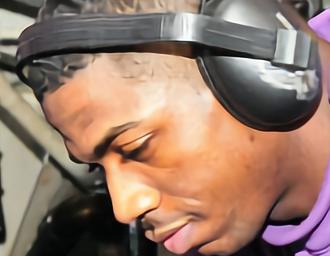}
   \put(0.14\linewidth, -1mm){
    \includegraphics[width=0.05\linewidth]{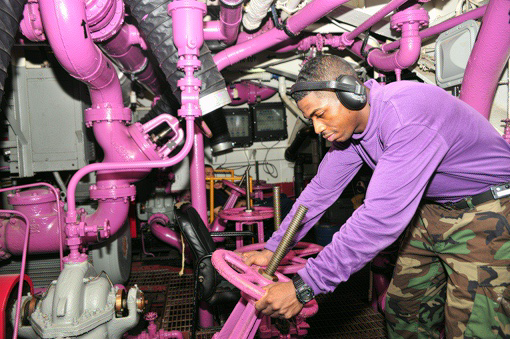}}
   \put(0.14\linewidth + 4.9mm, 2.3mm){
    \includegraphics[width=0.007\linewidth]{figures/section4_experiment/Visual_Comparison/bbox.png}}
\end{overpic}&
\begin{overpic}
    [width=0.19\linewidth]{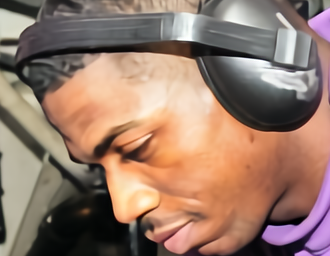}
   \put(0.14\linewidth, -1mm){
    \includegraphics[width=0.05\linewidth]{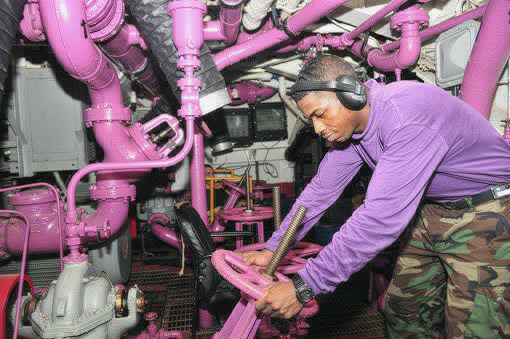}}
   \put(0.14\linewidth + 4.9mm, 2.3mm){
    \includegraphics[width=0.007\linewidth]{figures/section4_experiment/Visual_Comparison/bbox.png}}
\end{overpic}&
\begin{overpic}
    [width=0.19\linewidth]{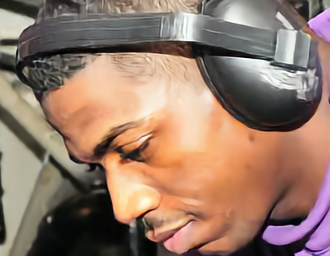}
   \put(0.14\linewidth, -1mm){
    \includegraphics[width=0.05\linewidth]{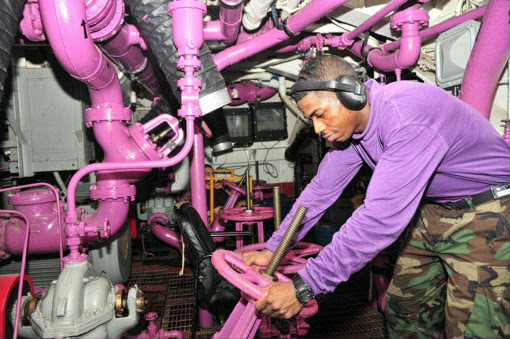}}
   \put(0.14\linewidth + 4.9mm, 2.3mm){
    \includegraphics[width=0.007\linewidth, trim=0 5 2 0, clip]{figures/section4_experiment/Visual_Comparison/bbox.png}}
\end{overpic}&
\begin{overpic}
   [width=0.19\linewidth]{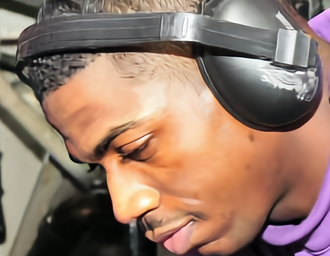}
   \put(0.14\linewidth, -1mm){
    \includegraphics[width=0.05\linewidth]{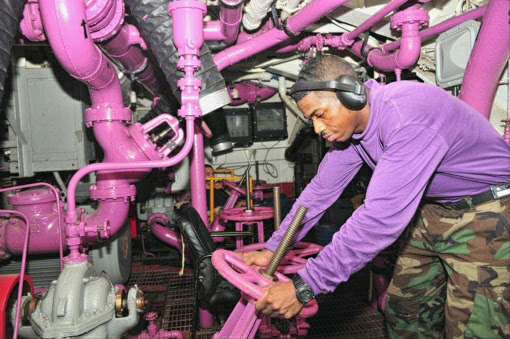}}
   \put(0.14\linewidth + 4.9mm, 2.3mm){
    \includegraphics[width=0.007\linewidth, trim=0 5 2 0, clip]{figures/section4_experiment/Visual_Comparison/bbox.png}}
\end{overpic}
\\
\footnotesize{bpp$\downarrow$~/~PSNR(RGB)$\uparrow$}& 
\footnotesize{0.350~/~25.45}&
\footnotesize{0.304~/~25.11}&
\footnotesize{\textbf{0.272~/~25.69}}&
\footnotesize{\textbf{0.271~/~26.03}}
\end{tabular}
\vspace{-0.5em}
\caption{\textbf{Reconstructed HR images and LR thumbnails by different methods on the DIV2K~\cite{agustsson2017ntire} validation dataset.} We crop the restored HR images to ease the comparison and visualize the LR counterparts at the bottom-right. The bpp is calculated on the whole image and the PSNR is evaluated on the cropped area of the reconstructed HR images. }
\vspace{-1em}
\label{fig:hr visual}
\end{figure*}

\section{Experiments}
\label{sec:experiments}
\subsection{Experimental Setup}
\noindent \textbf{Evaluation metrics}~
We evaluate our upscaling efficiency using running time and multiply-accumulation operations (MACs). 
We evaluate the quality of our reconstructed and embedding images using the PSNR on RGB channels. 

Rescaling methods are developed to embed an HR image into an LR one to reduce file size. However, there is no existing comprehensive evaluation of bitrate for these methods. Following previous compression approaches~\cite{balle2017end,balle2018variational,xie2021enhanced}, we evaluate the RD performance with the HR reconstruction PSNR against bits-per-pixel.
Instead of estimating the bitrate from the entropy encoder (Eqn.~\eqref{equ:entropy}) during training, 
we use the real file size of JPEG thumbnails for evaluating the bitrate:
\begin{eqnarray}
    {\rm bpp} = \mathbb{E}_{x \sim p_x}[\frac{\textbf{file size}}{H \times W}].    \label{equ:bpp_test}
\end{eqnarray}
Since there exists a trade-off between image bitrate and fidelity, we visualize the rate-distortion curve of different models during our ablation study.
\noindent \textbf{Datasets}~
We train our model with the widely-used DIV2K~\cite{agustsson2017ntire} image dataset, which contains 800 2K-resolution images in the training set and 100 more in the validation set. From the perspective of rescaling, we test our model on 4 conventional datasets: the Set5~\cite{bevilacqua2012low}, Set14~\cite{zeyde2010single}, BSD100~\cite{martin2001database}, and Urban100~\cite{huang2015single}. To reveal our real-world performance, we collect first 20 images above 6K resolution (ranked by their bpp) from MIT-Adobe FiveK~\cite{bychkovsky2011learning} as another test set FiveK-6k.
In addition, we evaluate the RD performance on Kodak dataset~\cite{KodakDataset}, which is widely adopted in image compression researches~\cite{begaint2020compressai}.
The details of the training strategy are provided in the supplement.

\subsection{Compare with Baselines}

We consider three categories of reconstruction methods as our baselines: (1) 
downscaling with Bicubic interpolation, compressing with standard JPEG codec, and upscaling with state-of-the-art SR models~\cite{lim2017enhanced,liang2021swinir}; 
(2) autoencoder using standard JPEG transmission format~\cite{jiang2018end}; 
(3) rescaling using symmetric invertible neural network~\cite{xiao2020invertible,liang2021hierarchical} and compressing the LR image using standard JPEG codec. 
For fair evaluation, we improve the RD performance of baselines using standard JPEG compression.
We retrain SR models~\cite{lim2017enhanced,liang2021swinir} on LR JPEG compressed images to restore HR images. Besides, we retrain INN baselines~\cite{xiao2020invertible,liang2021hierarchical} and encoder-decoder baselines~\cite{jiang2018end} with a differentiable JPEG module~\cite{shin2017jpeg}. 
To compare our RD performance against baselines, we constrain their bpp on Kodak dataset to be around 0.3 by adjusting the quality factor of JPEG compression in all baselines.
In our supplement, we provide an additional comparison with image compression (\eg, the original JPEG) and rescaling baselines using their original transmission format (\eg, lossless PNG), where our advantage is even larger.
\label{sec:evaluation}

\noindent \textbf{Upscaling efficiency and HR fidelity}~
Table~\ref{table:efficiency and fidelity} presents the upscaling efficiency of all methods and reconstruction fidelity at around 0.3 bpp.
We measure the running time and GMacs of upscaling a 960 $\times$ 540 resolution LR image to a 3840 $\times$ 2160 HR image on an Nvidia RTX 3090 GPU. 
We use PyTorch implementation with 16 bits floating-point precision for a fair comparison. As shown in Table~\ref{table:efficiency and fidelity}, ``Ours'' model only costs $3.1\%$ of the GMacs and $3.7\%$ of upscaling time compared to HCFlow~\cite{liang2021hierarchical}. We still improve the reconstructed RGB PSNR by 0.61 dB on BSD100 test set with a significantly lower computation.

In addition, we train the ``Ours-full'' model with a larger decoder. ``Ours-full'' model achieves higher PSNR that outperforms HCFlow~\cite{liang2021hierarchical} and IRN~\cite{xiao2020invertible} for more than 1.16 dB on Urban100 test set with only $24.2\%$ of upscaling time.
Fig.~\ref{fig:hr visual} provides the visual comparison of restored images. 
From left to right, we show perceptual differences from ground truth, $4\times$ rescaling results of the baselines~\cite{xiao2020invertible,liang2021hierarchical} with JPEG,
and our framework. For the first row, our models restore more textures of the glass and the pot lid. Similarly, among all reconstructions in the second row, only our model can recover the details on hair and earmuffs.
Besides, the JPEG compression breaks the invertibility of the INN-based rescaling methods~\cite{liang2021hierarchical,xiao2020invertible},
they failed to restore sharp and accurate high-frequency textures in the HR images.

\begin{figure}
\vspace{-2mm}
\centering
\begin{tabular}{@{}c@{\hspace{1mm}}c@{\hspace{1mm}}c@{\hspace{1mm}}c@{\hspace{1mm}}c@{}}
\scriptsize{Bicubic}& 
\scriptsize{IRN\cite{xiao2020invertible}$4\times$}&
\scriptsize{HCFlow\cite{liang2021hierarchical}$4\times$}&
\scriptsize{Ours}&
\scriptsize{Ours-full}
\vspace{-2mm}
\\
\scriptsize{$4\times$}& 
\scriptsize{\& JPEG q=96}&
\scriptsize{\& JPEG q=90}&
\scriptsize{$4\times$}&
\scriptsize{$4\times$}
\\
\begin{overpic}
    [width=0.19\linewidth]{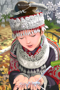}
    \put(0.06\linewidth, 0.06\linewidth){
    \includegraphics[width=0.07\linewidth, trim=0 0 0 0, clip]{figures/section4_experiment/Visual_Comparison/bbox.png}}
\end{overpic}&
\begin{overpic}
    [width=0.19\linewidth]{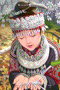}
    \put(0.06\linewidth, 0.06\linewidth){
    \includegraphics[width=0.07\linewidth, trim=0 0 0 0, clip]{figures/section4_experiment/Visual_Comparison/bbox.png}}
\end{overpic}&
\begin{overpic}
    [width=0.19\linewidth]{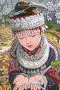}
    \put(0.06\linewidth, 0.06\linewidth){
    \includegraphics[width=0.07\linewidth, trim=0 0 0 0, clip]{figures/section4_experiment/Visual_Comparison/bbox.png}}
\end{overpic}&
\begin{overpic}
    [width=0.19\linewidth]{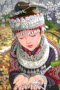}
    \put(0.06\linewidth, 0.06\linewidth){
    \includegraphics[width=0.07\linewidth, trim=0 0 0 0, clip]{figures/section4_experiment/Visual_Comparison/bbox.png}}
\end{overpic}&
\begin{overpic}
    [width=0.19\linewidth]{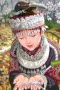}
    \put(0.06\linewidth, 0.06\linewidth){
    \includegraphics[width=0.07\linewidth, trim=0 0 0 0, clip]{figures/section4_experiment/Visual_Comparison/bbox.png}}
\end{overpic}
\\
\includegraphics[width=0.19\linewidth, trim=24 20 19 55, clip]{figures/section4_experiment/Visual_Comparison/4xlrs/comic_LR.png}&
\includegraphics[width=0.19\linewidth, trim=24 20 19 55, clip]{figures/section4_experiment/Visual_Comparison/4xlrs/comic_IRN.png}&
\includegraphics[width=0.19\linewidth, trim=24 20 19 55, clip]{figures/section4_experiment/Visual_Comparison/4xlrs/comic_HCFlow.jpg}&
\includegraphics[width=0.19\linewidth, trim=24 20 19 55, clip]{figures/section4_experiment/Visual_Comparison/4xlrs/comic_Fast.jpg}&
\includegraphics[width=0.19\linewidth, trim=24 20 19 55, clip]{figures/section4_experiment/Visual_Comparison/4xlrs/comic_Full.png}
\\
\scriptsize{bpp$\downarrow$ /PSNR$\uparrow$}& 
\scriptsize{0.510~/~26.67}&
\scriptsize{0.468~/~17.47}&
\scriptsize{\textbf{0.430~/~27.99}}&
\scriptsize{\textbf{0.446~/~27.90}}
\end{tabular}
\vspace{-0.5em}
\caption{\textbf{Downscaled LR thumbnails by different methods on Set14 image \textit{comic}.} With a similar target bpp, our model introduces least artifacts in the thumbnail in comparison to baselines.}
\label{fig:lr_psnr}
\end{figure}

\begin{table}[t]

\centering
\resizebox{1\linewidth}{!}{
\begin{tabular}{@{}l@{\hspace{1mm}}c@{\hspace{1mm}}*{6}{c@{\hspace{1mm}}}}

\toprule
Method & \multicolumn{6}{c}{LR thumbnail PSNR$\uparrow$} \\
\cmidrule(l{0mm}r{1mm}){1-1} 
\cmidrule(l{1mm}r{1mm}){2-8}
Down \& Degradation & Kodak & Set5 &  Set14  &  BSD100  &  Urb100 &  DIV2K & FiveK-6k \\
\midrule

Bicubic \& JPEG & 37.72 & 34.41 & 35.79 & 37.43 & 36.64 & 37.14 & 35.14 \\

IRN~\cite{xiao2020invertible} \& JPEG & 30.95  & 30.00  & 27.23  & 26.91  & 25.72  & 29.54 & 31.25 \\

ComCNN \& RecCNN \cite{jiang2018end} & 28.00 & 26.76 & 26.47 & 27.47 & 25.57 & 28.15 & 28.99
\\
HCFlow~\cite{liang2021hierarchical} \& JPEG & 19.88  & 20.08  & 19.42  & 19.65  & 18.96  & 20.52 & 20.31 \\
\midrule
Ours-full & 33.21  & 31.86  & 31.76  & 32.44  & 31.01  & 33.32 & 33.99\\
Ours & 33.55 & 31.96 & 31.93 & 32.90 & 31.16 & 33.62 & 34.24 \\


\bottomrule
\end{tabular}
}
\vspace{-1mm}
\caption{Quantitative evaluation of the $4\times$ downsampled LR thumbnails by different methods. The target bitrate is around 0.3 bpp on Kodak~\cite{KodakDataset} for all methods, and we take Bicubic LR as the ground truth. Our thumbnail preserves visual contents better.}
\label{table:LR}
\vspace{-1em}
\end{table}

\begin{figure}[ht]
    \centering
    \begin{tabular}{@{}c@{\hspace{2mm}}c@{\hspace{2mm}}c@{}}
        \scriptsize{(a) Encoder}  & \scriptsize{(b) Decoder}
        \\
        \includegraphics[width=0.48\linewidth]{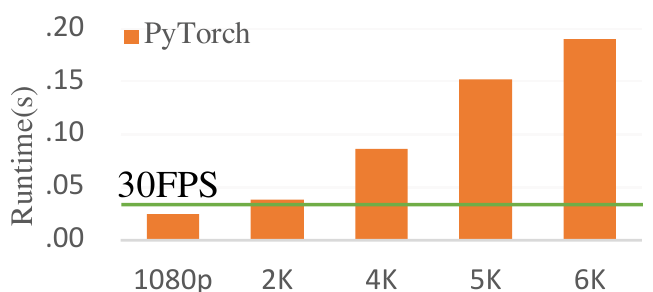}&
        \includegraphics[width=0.52\linewidth]{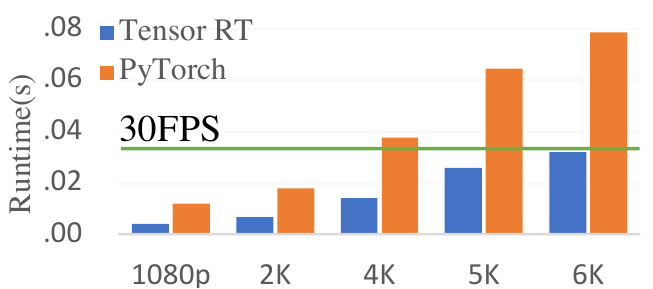}
    \end{tabular}
    \vspace{-1em}
    \caption{\textbf{Model runtime.} We profile the $4\times$ encoder and decoder at different target resolution in half-precision mode. Especially, we convert our decoder from PyTorch to TensorRT for further inference time reduction. }
    \label{fig:trt_runtime}
    \vspace{-1em}
\end{figure}

\noindent \textbf{LR qualitative evaluation}~
The visual quality of the LR thumbnails is also important because users preview them directly. Table~\ref{table:LR} presents the average LR PSNR of corresponding models, we have the best LR PSNR against INN-based methods. 
In Fig.~\ref{fig:lr_psnr}, we qualitatively compare the visual quality of the thumbnails encoded by rescaling methods. Our method generates LR thumbnails with significantly fewer artifacts without compromising much HR reconstruction quality, please see more analysis in Sec.~\ref{sec:ablation}.

\subsection{Real-time Inference on 6K Images}
Fig.~\ref{fig:trt_runtime} shows our evaluation of the downscaling and upscaling runtime of our framework at multiple resolutions. Noted that we conduct the profiling in half-precision (FP16) mode. Besides, we also converted the trained decoder model from PyTorch to FP16 TensorRT model with torch2trt~\cite{torch2trt} API to further reduce the upscaling time without performance drop in HR reconstruction.
As shown in Fig.~\ref{fig:trt_runtime}(b), our efficient decoder can upscale an LR thumbnail by $4\times$ in \textbf{real time} at 4K~($3840\times2160$) \textbf{70.8} FPS, 5K~($5120\times2880$) \textbf{38.8} FPS, or 6K~($5760\times3240$) \textbf{31.2} FPS on an RTX 3090 GPU.







\begin{figure*}[t]
\centering
\vspace{-0.5em}
\begin{tabular}{@{}c@{\hspace{1.2mm}}c@{\hspace{1.2mm}}c@{\hspace{1.2mm}}c@{\hspace{1.2mm}}c@{}}
\scriptsize{kodim04 / kodim09}&
\scriptsize{(a) Ours with Fixed JPEG tables, q=85} &
\scriptsize{(b) Ours with Optimized tables}& 
\scriptsize{(c) Ours with QPM kodim04}&
\scriptsize{(d) Ours with QPM kodim09} 
\\
\includegraphics[height=0.1\linewidth]{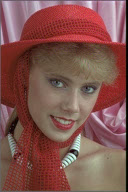} \includegraphics[height=0.1\linewidth]{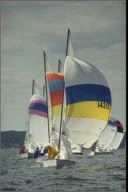}&
\includegraphics[width=0.1\linewidth]{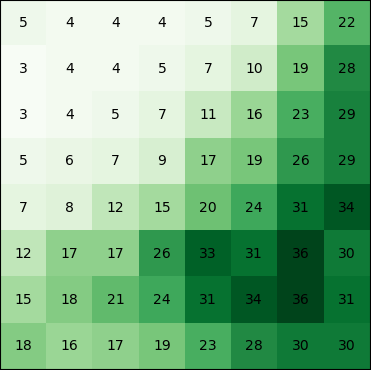} \includegraphics[width=0.1\linewidth]{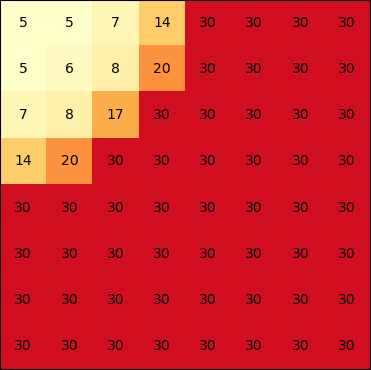}&
\includegraphics[width=0.1\linewidth]{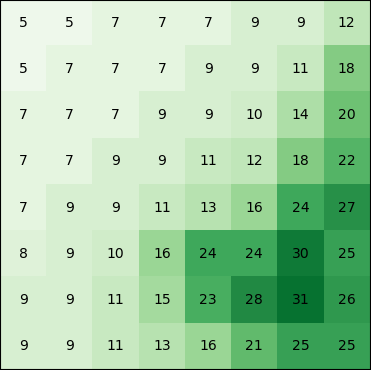} \includegraphics[width=0.1\linewidth]{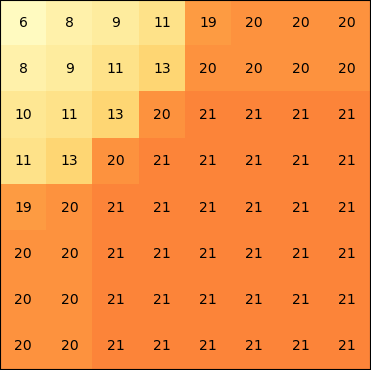}&
\includegraphics[width=0.1\linewidth]{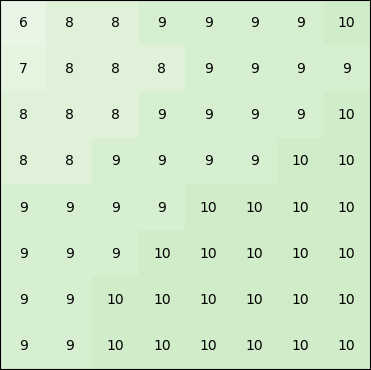} \includegraphics[width=0.1\linewidth]{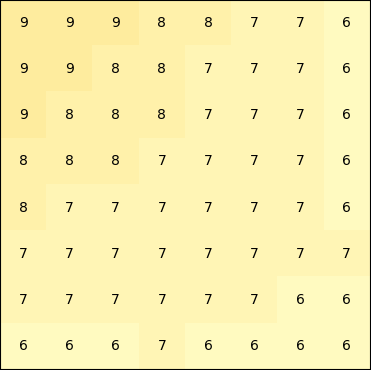}&
\includegraphics[width=0.1\linewidth]{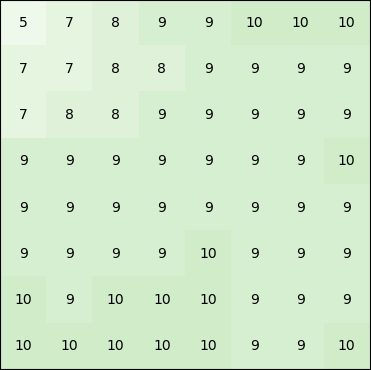} \includegraphics[width=0.1\linewidth]{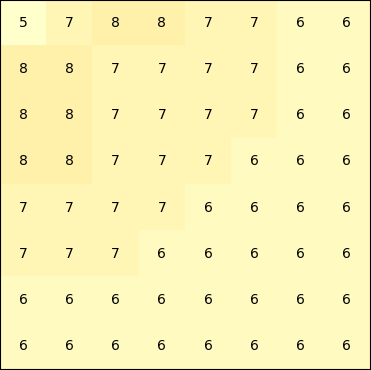}
\\
\scriptsize{bpp$\downarrow$ / LR$\uparrow$ / HR$\uparrow$: \textit{kodim09}}&
\scriptsize{0.263 / 26.91 / 30.54}&
\scriptsize{0.289 / 29.10 / 30.93}&
-&
\textbf{\scriptsize{0.246 / 31.03 / 34.34}}
\\
\scriptsize{bpp$\downarrow$ / LR$\uparrow$ / HR$\uparrow$: \textit{kodim04}}&
\scriptsize{0.251 / 26.96 / 31.35}&
\scriptsize{0.276 / 29.17 / 31.71}&
\textbf{\scriptsize{0.240 / 34.38 / 32.03}}&
-
\vspace{-0.8em}
\end{tabular}
\caption{\textbf{Quantization tables on Kodak~\cite{KodakDataset} images.} We visualize the quantization table $Q_L$ (the green table) and $Q_C$ (the orange table) for \textit{kodim04} and \textit{kodim09} of different quantization approaches. The model trained with QPM achieves the best RD performance from every aspect. For more analysis, please refer to Sec.~\ref{sec:ablation} in our paper.}
\label{fig:quant_table}
\vspace{-1em}
\end{figure*}

\subsection{Extension for Optimization-based Rescaling}
Since the ground truth HR images are available for downscaling during the test stage, we may further optimize our encoder $E$, QPM, and Entropy model, while fixing the pretrained decoder $D$ and feature extractor $f$ on the user's device. As shown in the Table~\ref{table:optimization}, optimization-based rescaling improves the restoration PSNR 0.22dB with a lower bitrate on Set14 dataset.

\begin{table}[h]
\centering
\resizebox{1\linewidth}{!}{
\begin{tabular}{@{}l@{\hspace{1mm}}c@{\hspace{3mm}}*{5}{c@{\hspace{3mm}}}}

\toprule


Method &
\multicolumn{1}{c}{Optimization} &
\multicolumn{3}{c}{bpp$\downarrow$ / HR PSNR$\uparrow$}
\\
\cmidrule(l{0mm}r{1mm}){1-1}
\cmidrule(l{1mm}r{1mm}){2-2}
\cmidrule(l{1mm}r{1mm}){3-5}
Architecture & iteration & Kodak & Set5 & Set14\\
\midrule
\multirow{2}{*}{Ours} & 0 & 0.301 / 29.42 & 0.379 / 30.23 & 0.359 / 27.74 \\
~ & 100 & 0.307 / 29.55 & 0.377 / 30.36 & 0.347 / 27.96 \\

\bottomrule
\end{tabular}
}

\vspace{-1mm}
\caption{Quantitative evaluation for optimization-based rescaling.}
\vspace{-1em}
\label{table:optimization}
\end{table}

\section{Ablation Study}
In this section, we study the effectiveness of our proposed Quantization Prediction Module, designed training loss and architectures on ``Ours" model.
\label{sec:ablation}

\begin{figure}
    \centering
    \begin{tabular}{@{}c@{\hspace{0mm}}c@{}}
        \scriptsize{(a) RD curve of restored HR image}  & \scriptsize{(b) RD curve of LR thumbnail}
        \\
        \includegraphics[width=0.5\linewidth]{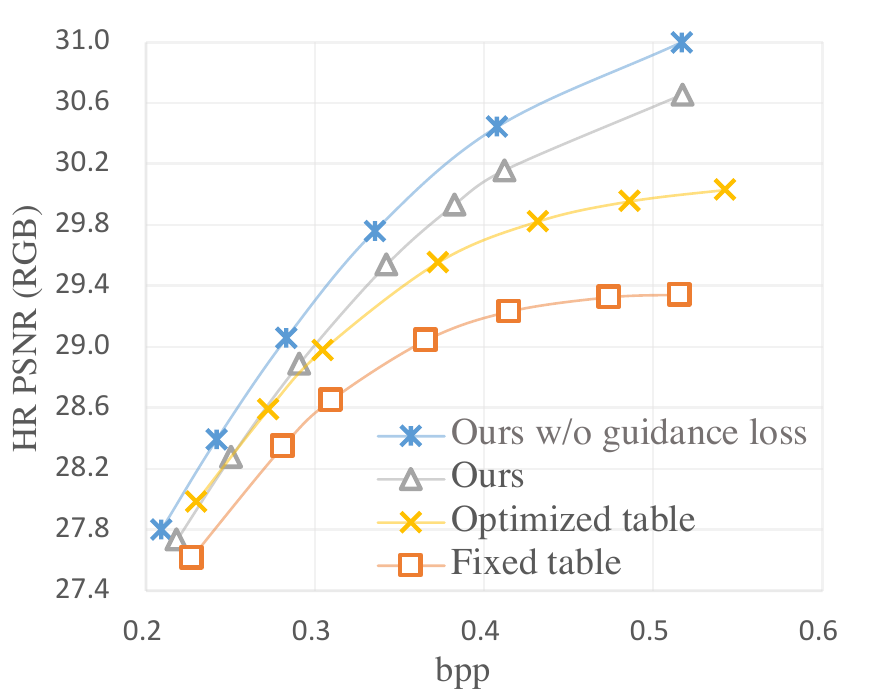}&
        \includegraphics[width=0.5\linewidth]{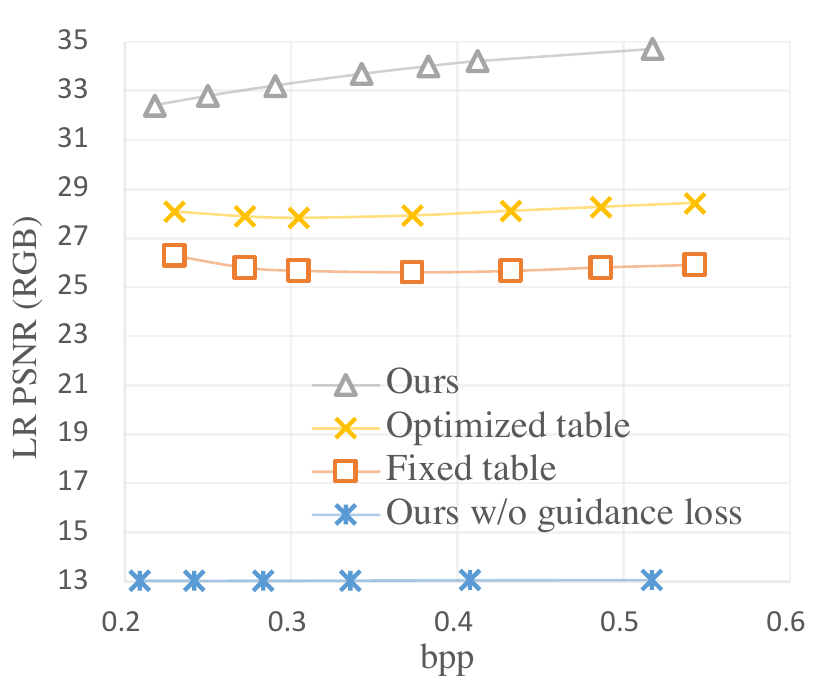}
    \end{tabular}
    \vspace{-1em}
    \caption{\textbf{QPM versus image-invariant quantization.} We first train our models with QPM, with a fixed JPEG table or with an optimized table, respectively. Then, we evaluate the at different target bitrate on Kodak~\cite{KodakDataset} dataset. \textbf{(a)} the RD curve on reconstructed HR image $\hat{x}$ and input $x$; \textbf{(b)} the RD curve on LR thumbnail $\hat{y}$ and the Bicubic downsampled LR $y_{ref}$.}
    \label{fig:abla_qpm}
    \vspace{-1.5em}
\end{figure}

\vspace{0.8em}
\noindent\textbf{Quantization prediction module}~
To examine the effectiveness of our QPM, in Fig.~\ref{fig:abla_qpm}, we quantitatively evaluate the RD performance on both the restored HR and the LR JPEG thumbnails of different quantization approaches. For ``fixed tables'' and ``optimized tables'', we initialize the quantization tables following the default JPEG. 
Particularly, for ``optimized tables,'' we also optimize the quantization tables at the training stage.
We illustrate the curve for each model by adjusting the global quality factor $q$ on the quantization tables $Q$ following JPEG as 
$Q' = Q \times q.$
The target bitrate goes lower when $q$ increases. 
In Fig.~\ref{fig:quant_table}, we also visualize the quantization tables of different settings.
We notice that compared to the corresponding value in the fixed or optimized table, the high-frequency quantization steps in the QPM predicted tables are much smaller and are image-specific, which may introduce less compression on the embedding pattern that is important for HR reconstruction. Consequently, compared to settings with image-invariant quantization, ``Ours" model achieves the best quality on the reconstructed $\hat{x}$ and introduces significantly fewer artifacts on the LR thumbnail $\hat{y}$, which also lowers the filesize of the thumbnails.
To further investigate the effectiveness of QPM, we also leverage QPM to improve the RD performance of standard JPEG (without the downscaling and upscaling). Please refer to our supplement for more details.

\vspace{0.8em}
\noindent\textbf{Guidance loss}~
In Fig.~\ref{fig:abla_qpm}, we also present the RD curve of the ``Ours w/o guidance loss" model trained with $\lambda_1 = 0$. 
At around 0.4 bpp, 
removing guidance loss
raises the HR PSNR by 0.09dB. However, the PSNR of LR thumbnail drops significantly from 34.1 dB to 13.06 dB. It is unacceptable for user-viewable thumbnails.

\vspace{0.8em}
\noindent\textbf{Encoder decoder architecture}~
As we study some variants of ``Ours" model, we find that $f$ improves the quality of both encoded LR thumbnail (0.29 dB) and restored HR image (0.10 dB). Also, adopting $f$ is more effective than simply increasing the decoder capacity.
Moreover, our study of the encoder capacity reveals that our framework does benefit from a larger encoder. More details are in our supplement.

\section{Conclusion}
\vspace{-0.1em}
In this paper, we propose a new HyperThumbnail framework that can perform real-time 6K image reconstruction from an LR JPEG thumbnail. We utilize an asymmetric encoder-decoder architecture where the encoder takes most of the computation while the decoder is relatively lightweight for real-time performance. A new quantization prediction module is proposed to optimize the RD performance for image rescaling, which is not studied in prior work. Our framework benefits image sharing and transfer in real-world latency-sensitive applications, such as cloud photo storage and retrieval. 

\noindent\textbf{Acknowledgement}
We express our sincere gratitude to our friends and domain experts, Junming Chen, Yue Wu, and Yu Wang for their invaluable contributions to the design of our project. Additionally, we extend our appreciation to Xiaogang Xu and Xilin Zhang for reviewing and revising the writing.

{\small
\bibliographystyle{ieee_fullname}
\bibliography{egbib}
}
\newpage
\clearpage
\newpage
\appendix

\section*{Summary}
\noindent This supplementary material is organized as follows.
\begin{itemize}
    \renewcommand{\labelitemi}{\textbullet}
        \item Section~\ref{sec:Implementation Details} introduces the implementation of our architecture and training details.
        \item Section~\ref{Sec:Additional Comparison Results} shows more comparison with previous work.
        \item Section~\ref{sec:Additional ablation} discusses more ablation studies of our designs.
\end{itemize}


\section{Implementation Details}
\label{sec:Implementation Details}
We implement our model in PyTorch and train on a single Nvidia RTX3090 GPU. In this section, we describe the details of our architecture and training settings.

\noindent\textbf{Network Architecture.}\quad
Table~\ref{table:encoder} and Table~\ref{table:decoder} respectively show details of our encoder and decoder. In Table~\ref{table:encoder}, we describe the architecture of our UNet-based encoder~\cite{ronneberger2015unet}. ``PixelUnshuffle 4x'' stands for the rearrangement of elements~\cite{shi2016real} which downsamples the HR image by a factor of 4. ``3×3, 64, LeakyReLU" denotes a 2d convolution operation of kernel size 3, output channel 64, followed by a LeakyReLU operation.
We use the implementation of Residual Dense Block from~\cite{wang2018esrgan}, where "ResidualDenseBlock-32" refers to a Residual Dense Block with minimum channel 32. 
We save the produced quantization tables and output coefficients into a single JPEG file using TorchJPEG~\cite{ehrlich2020quantization}.
Building blocks are shown in brackets, with the number of blocks stacked. Downsampling is performed at the beginning of the downsampling block using max pooling of stride 2. 
After the first convolution in the upsampling block, we upsample the decoder feature with Pixel Shuffling Operation. Then, skip connections with the encoder features are conducted.

\begin{table}[ht!]

\small
\centering
\renewcommand{\arraystretch}{1.0}
\resizebox{\linewidth}{!}{
\begin{tabular}{l@{\hspace{2mm}} l@{\hspace{2mm}} r}
\toprule
Stage & Building Block & Output Size\\
\midrule
Input Downsample
& 
{$\begin{array}{l}
\text{PixelUnshuffle 4x} \\
3 \times 3, 64, \\
\text{LeakyReLU} \\
\text{ResidualDenseBlock-32} \\
\end{array}
$ }  
& ${H/4} \times {W/4} \times 64 $ \\

\hline
\text{Downsampling Block1}
& 
{$\begin{array}{l}
{
\left[
\begin{array}{l}
3 \times 3, 128, \\
\text{ LeakyReLU}
\end{array}
\right] \times 2 
} \\
\text{ResidualDenseBlock-64}
\end{array}$}
& $H/8 \times W/8 \times 128$  \\ 

\hline
\text{Downsampling Block2}
& 
{$\begin{array}{l}
{\left[\begin{array}{l}
3 \times 3, 256, \\
\text{  LeakyReLU}\end{array}\right] \times 2
}  \\
\text{ResidualDenseBlock-128}
\end{array}$}
& $H/16 \times W/16 \times 256$  \\

\hline
\text{Upsampling Block1}
& 
{$\begin{array}{l}
3 \times 3, 512 \\
{\left[\begin{array}{l}
3 \times 3, 128, \\
\text{ LeakyReLU} \\
\end{array}\right] \times 2}  \\
\text{ResidualDenseBlock-128}
\end{array}$}
& $H/16 \times W/16 \times 128$  \\ 

\hline
\text{Upsampling Block2}
& 
{$\begin{array}{l}
3 \times 3, 256 \\
{\left[\begin{array}{l}
3 \times 3, 64, \\
\text{ LeakyReLU} \\
\end{array}\right] \times 2}  \\
\text{ResidualDenseBlock-64}
\end{array}$}
& $H/8 \times W/8 \times 64$  \\ 

\hline
Output layer &
$\begin{array}{l}
    3 \times3, 3 \\
\end{array}{}$
 & $H/4 \times W/4 \times 3$  \\ 

\bottomrule
\end{tabular}
}
\caption{\textbf{Architectures of our encoder.}
}
\vspace{-3mm}
\label{table:encoder}
\end{table}

In Table~\ref{table:decoder}, we show the details of our efficient decoder, which is developed based on EDSR~\cite{lim2017enhanced}. 
We extract features $f(\widehat{C})\in \mathbb{R}^{24 \times \frac{H}{s}\times \frac{W}{s}}$ and concatenate $f(\widehat{C})$ with the RGB image $\hat{y}$. The concatenated features are fed into the decoder to reconstruct the HR image $\hat{x}$:
\begin{eqnarray}
\hat{x} &=& D(\hat{y} \oplus f(\widehat{C})),
\end{eqnarray}
where $\oplus$ is the concatenation operator along the channel dimension.
\begin{table}[ht]

\centering
\renewcommand{\arraystretch}{1.1}
\resizebox{1.0\linewidth}{!}{
\begin{tabular}{l l r}
\toprule
Frequency Feature Extractor $f$ & Building Block & Output Size\\

\midrule
\text{Input Convolution}
& {
$\begin{array}{l}
3 \times 3, 24 \\
\end{array} 
$ }  
& $H/32 \times W/32 \times 24 $ \\ 

\midrule
\text{Residual Convolution Block}
& {$\left[\begin{array}{l}
3 \times 3, 24,\\
\text{ReLU},\\
3 \times 3, 24, \\
\end{array}\right] \times 16$ }  & $H/32 \times W/32 \times 24$  \\ 

\midrule
Output Convolution &
$\left[\begin{array}{l}
3 \times 3, 96,\\
\text{PixelShuffle 2x} \\
\end{array}\right] \times 3
$ 
& $H/4 \times W/4 \times 24$  \\ 
\bottomrule
\end{tabular}
}
\vspace{1em}
\\

\vspace{1em}
\resizebox{1.0\linewidth}{!}{
\begin{tabular}{l l r}
\toprule
Decoder-full & Building Block & Output Size\\
\midrule
\text{Input Convolution}
& {
$\begin{array}{l}
3 \times 3, 24 \\
\end{array}
$ }  
& $H/4 \times W/4 \times 24 $ \\ 

\hline
\text{RRDB Blocks}
& {$
\left[\begin{array}{l}
\text{ResidualDenseBlock-32}, \\
\text{ResidualDenseBlock-32},\\
\text{ResidualDenseBlock-32}, \\
\end{array}\right] \times 12
$

}  & $H/4 \times W/4 \times 24$  \\ 

\hline
Output Convolution & 
$\left[\begin{array}{l}
3 \times3, 96,\\
\text{PixelShuffle 2x}\\
\end{array}\right] \times 2
$
& $H \times W \times 3$  \\ 
\bottomrule
\end{tabular}
}

\caption{\textbf{Architectures of our efficient decoder.}}
\vspace{-1em}
\label{table:decoder}
\end{table}

\noindent\textbf{Training with Pixel Loss.}\quad
The model is trained with batch size 16 and patch size $256 \times 256$ in each iteration. The initial learning rate is 2e-4. The learning rate is decayed by $0.75$ for every $100,000$ iterations.

\noindent\textbf{Test-time Fine-tuning during Downscaling.}\quad
During downscaling stage, we optimize the pre-trained encoder with a fixed pretrained decoder. In the optimization during downscaling, we use full-resolution test images without augmentation as batch size 1 to accelerate optimization. 
For each image in the test set, we optimize the encoder and QPM for 100 iterations with a learning rate of $2.0\times10^{-4}$. 

\section{Additional Comparison Results}
\label{Sec:Additional Comparison Results}
\subsection{Comparison with Compression+JPEG}

The key difference between our rescaling framework with learned image compression ~\cite{balle2018variational, balle2017end, mentzer2020high} is that our HyperThumbnail provides an instant preview that is compatible with existing JPEG codec. However, learned image compression typically requires GPU for decompression using neural networks. For users of learned compression, one practical solution to support instant preview is saving a low-resolution JPEG image as a thumbnail besides compressed bitstream, which we refer to as ``Compression+JPEG'' framework.

Our rescaling framework has two advantages over the above ``Compression+JPEG'' solution. (a) First, we embed the high-frequency information into a compact single JPEG file that is easy to deliver. In contrast,  the ``Compression+JPEG'' framework requires two different file formats for preview and compressed bitstreams, which is inconvenient for storage and transmission.
(b) Secondly, as evaluated in Table~\ref{table:Compresssion+JPEG}, it takes considerable storage for standard JPEG~\cite{wallace1991jpeg} thumbnails to have similar fidelity as our encoded LR thumbnails. We choose Hyperprior~\cite{balle2018variational} and HIFIC~\cite{mentzer2020high}, two state-of-the-art compression methods with a similar running time as ours to build ``Compression+JPEG'' baseline. Because of information redundency in the bitstream and the  JPEG file, ``Compression+JPEG'' framework takes more storage to achieve comparable LR PSNR and HR PSNR with our result. In summary, our HyperThumbnail provides a compact and succinct representation to support both instant preview and high-frequency reconstruction.

\begin{table}[t]
\centering
\resizebox{1\linewidth}{!}{
\begin{tabular}{@{}l@{\hspace{5mm}}*{3}c@{\hspace{2mm}}c@{\hspace{2mm}}}

\toprule
Method & \multicolumn{2}{c}{Bpp of File Format} &
\multicolumn{2}{c}{Bitrate$\downarrow$-Distortion$\uparrow$ Kodak}
\\
\cmidrule(l{0mm}r{1mm}){1-1}
\cmidrule(l{1mm}r{1mm}){2-3}
\cmidrule(l{1mm}r{1mm}){4-5}

Architecture & Bitstream & JPEG &  Sum of bpp & LR PSNR / HR PSNR    \\
\midrule
Hyperprior~\cite{balle2018variational}+JPEG & 0.214 & 0.148 & 0.51 & 33.41 / 29.22   \\
HIFIC~\cite{mentzer2020high}+JPEG & 0.172 & 0.148 & 0.32 & 33.41 / 29.35   \\
Ours & - & 0.299 &\textbf{0.30} & \textbf{33.55} / \textbf{29.42}   \\

\bottomrule
\end{tabular}
}
\vspace{-1mm}
\caption{Comparison of our HyperThumbnail framework against learned compression with JPEG thumbnail. In additional baseline, we provide a JPEG thumbnail besides learned compression, and take the sum of bitstream size and JPEG size to calculate the final bpp. Our framework has better rate-distortion performance than ``Compression+JPEG'' baseline.}
\label{table:Compresssion+JPEG}
\end{table}

\subsection{Quantitative comparison with JPEG}

\begin{figure}[t!]
    \centering
    \includegraphics[width=1.0\linewidth]{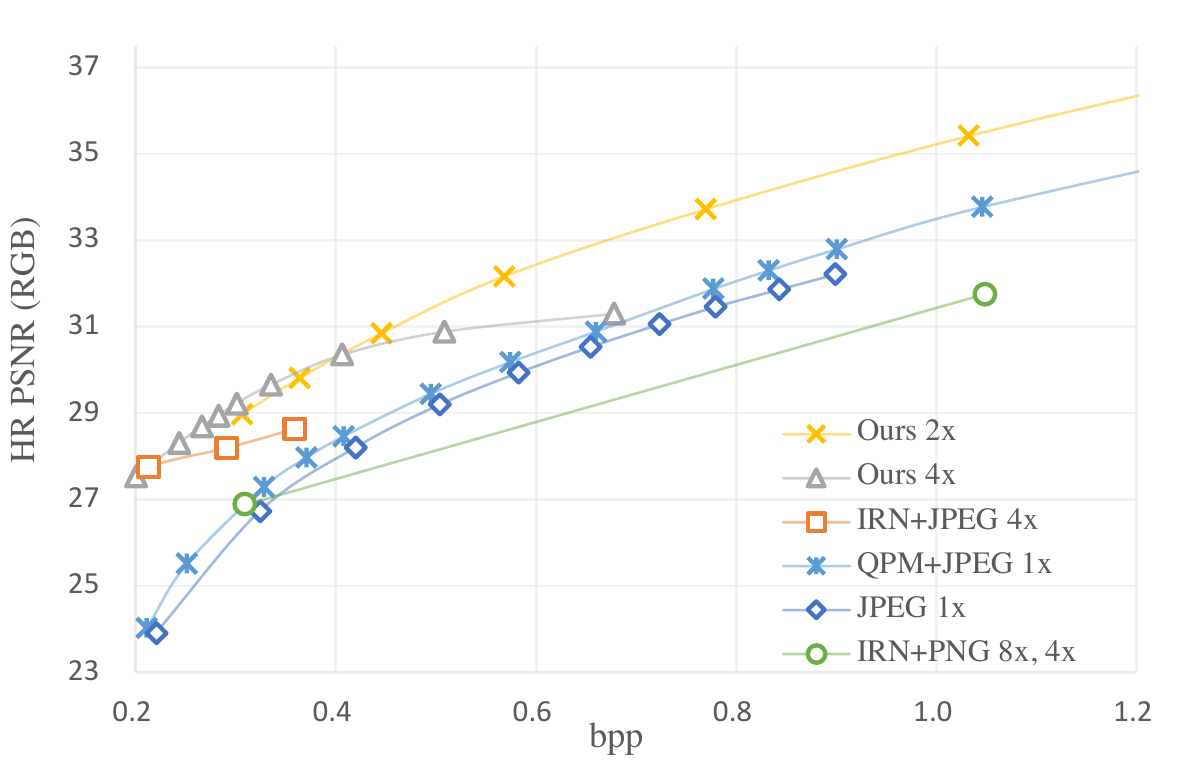} 
    \caption{\textbf{The rate-HR-distortion curve on Kodak~\cite{KodakDataset} dataset.} Our method ($s=2, 4$) outperforms JPEG, IRN~\cite{xiao2020invertible} in the RD performance. For the `QPM + JPEG' curve, where $s=1$, we follow the standard JPEG algorithm and adopt QPM module as a plugin for table prediction.}
    \label{fig:rate_distortion}
\end{figure}

In Figure.~\ref{fig:rate_distortion}, we provide an additional comparison of our rate-HR-distortion performance with baselines. 
Previous rescaling methods such as IRN~\cite{xiao2020invertible} in PNG format with different rescaling scale (``IRN+PNG $8\times$,$4\times$'') is even worse than ``JPEG''~\cite{wallace1991jpeg}.
``IRN+JPEG $4\times$'' shows that JPEG format with different quality factors boosts the rescaling methods. In contrast, our method is much better than the above three baselines, thanks to our image-specific quantization design. 

Another interesting extension of our work is to use QPM as a plug-in to improve the performance of traditional JPEG compression, which is shown by ``QPM + JPEG'' curve. We set the rescaling factor as $s=1$, remove our encoder and decoder, and only train our QPM Module as a compression method. An improvement of 0.5dB is observed at most bitrate constraints.

Note that traditional image compression codec, such as JPEG, does not produce an LR embedding as rescaling methods. Thus, their results are only for reference.

\subsection{Decoding efficiency comparison with AVIF and JPEGXL.}

With no available GPU implementation, we test the decoding efficiency of AVIF \textbf{(344.8 ms)} and JPEG-XL \textbf{(257.9 ms)} on an Intel Xeon Gold 5218 server CPU at 4K resolution and 0.3 bpp. 
In comparison, our decoder \textbf{(14.1 ms)} is much faster with GPU acceleration. 
According to a survey~\cite{JPEGUsageStatistics}, the usage statistics of JPEG \textbf{(77.8\%)} is much higher than AVIF \textbf{(<0.1\%)}. 
Meanwhile, JPEG-XL will soon be deprecated by Chrome, and some websites (e.g., Twitter and Shopee) use JPEG as the only lossy image file format.
Meanwhile, our framework can be integrated into most apps (e.g., Chrome and WhatsApp) without building extra support for transmission and previewing, which is more practical and useful.

\begin{table*}[ht!]
\centering
\resizebox{0.7\linewidth}{!}{
\begin{tabular}{@{}l@{\hspace{1mm}}c@{\hspace{1mm}}*{5}{c@{\hspace{1mm}}}}

\toprule
Method & \multicolumn{2}{c}{Bitrate$\downarrow$-Distortion$\uparrow$ Kodak}&
 \multicolumn{1}{c}{ Time(ms)$\downarrow$} &
 \multicolumn{3}{c}{ LR PSNR $\uparrow$ / HR PSNR$\uparrow$} \\
\cmidrule(l{0mm}r{1mm}){1-1}
\cmidrule(l{1mm}r{1mm}){2-3}
\cmidrule(l{1mm}r{1mm}){4-4}
\cmidrule(l{1mm}r{1mm}){5-7}
Architecture & bpp & LR PSNR / HR PSNR &  \textcolor{magenta}{Down} / \textcolor{cyan}{Up}  &  BSD100  &  Urb100 &  DIV2K \\
\midrule

Ours w/o $f$ & 0.30 & 33.26 / 29.32 & \textcolor{magenta}{86.2} / \textcolor{cyan}{32.3} & 32.56 / 27.57~~ & 30.90 / 26.46~~ & 33.40 / 30.00 \\

Ours- w/o $f$-b22 & 0.30 & 33.24 / 29.27 & \textcolor{magenta}{86.2} / \textcolor{cyan}{38.6} & 32.51 / 27.63~~ & 30.89 / 26.69~~ & 33.48 / 30.09 \\

Ours & 0.30 & 33.55 / 29.42 & \textcolor{magenta}{86.2} / \textcolor{cyan}{37.8} & 32.90 / 27.66~~ & 31.16 / 26.62~~ & 33.62 / 30.15 \\


\midrule

Ours enc-48 & 0.30 & 33.51 / 29.17  & \textcolor{magenta}{63.7} / \textcolor{cyan}{37.8} & 32.88 / 27.58~~ & 31.21 / 26.51~~ & 33.62 / 30.05 \\

Ours enc-96 & 0.30 & 33.52 / 29.29 & \textcolor{magenta}{183.5} / \textcolor{cyan}{37.8} & 32.88 / 27.66~~ & 31.22 / 26.66~~ & 33.62 / 30.15 \\

\bottomrule
\end{tabular}
}
\vspace{-1mm}
\caption{Ablation study of our encoder-decoder architectures on the \textcolor{magenta}{downsampling} / \textcolor{cyan}{upsampling} time and the PSNR of reconstructed HR image / LR thumbnail.}
\label{table:abla_arch}
\end{table*}

\section{Additional Ablation Study}
\label{sec:Additional ablation}

\begin{figure}
\centering
\begin{tabular}{@{}c@{\hspace{9mm}}c@{\hspace{9mm}}c@{}}
\scriptsize{(a) Ground truth} &
\scriptsize{(b) Ours $\lambda_1=0.6$}& 
\scriptsize{(c) Ours $\lambda_1=0$}
\\
\begin{overpic}
    [width=0.20\linewidth, trim=156 460 156 116, clip]{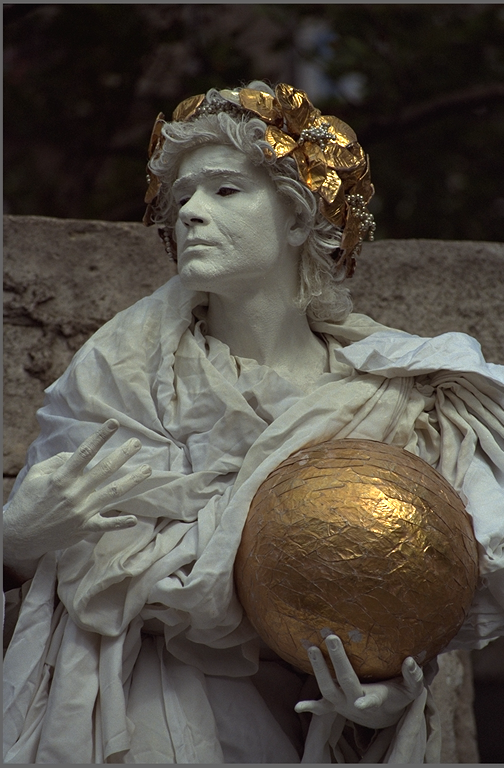}
    \put(0.15\linewidth,-0.02\linewidth){
    \includegraphics[width=0.10\linewidth, trim=39 115 39 29, clip]{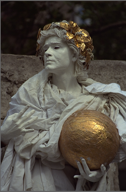}
    }
\end{overpic}&
\begin{overpic}
    [width=0.20\linewidth, trim=160 456 160 120, clip]{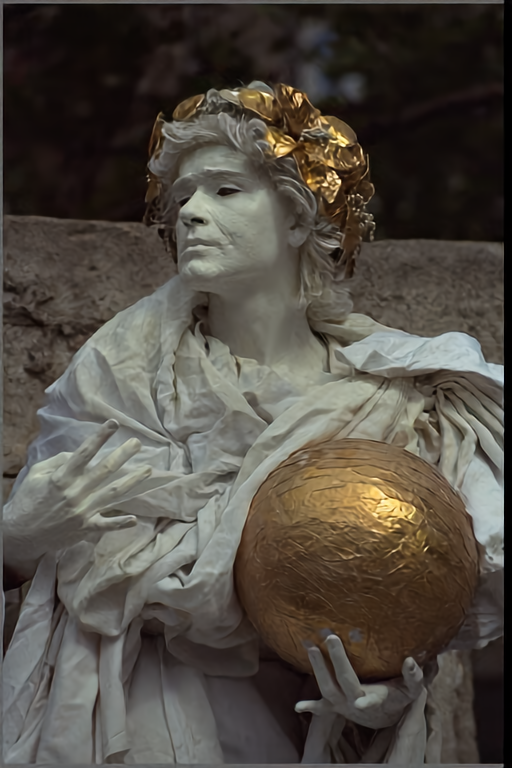}
    \put(0.15\linewidth,-0.02\linewidth){
    \includegraphics[width=0.10\linewidth, trim=40 114 40 30, clip]{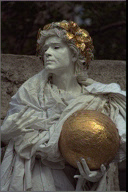}
    }
\end{overpic}&
\begin{overpic}
    [width=0.20\linewidth, trim=160 456 160 120, clip]{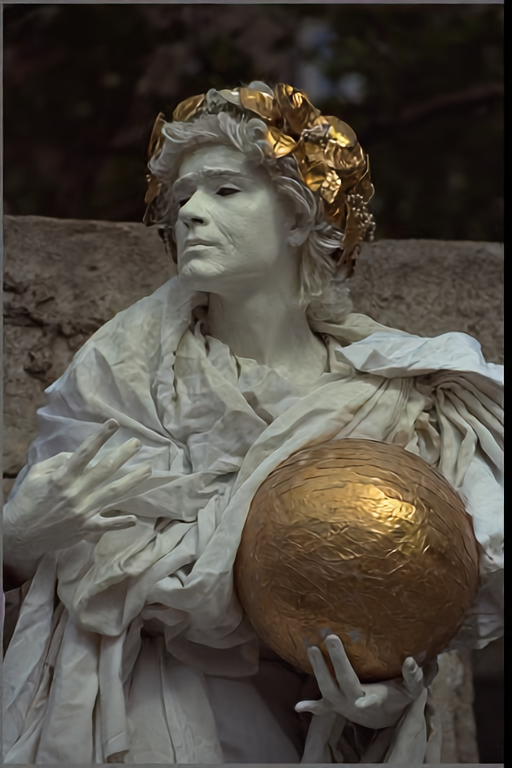}
    \put(0.15\linewidth,-0.02\linewidth){
    \includegraphics[width=0.10\linewidth, trim=40 114 40 30, clip]{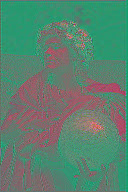}
    }
\end{overpic}
\end{tabular}
\vspace{-0.5em}
\caption{\textbf{guidance loss ablation on Kodak~\cite{KodakDataset} image \textit{kodim17}.} We visualize the HR images with their LR counterparts at the bottom-right. (b) (c) are produced by $4\times$ HyperThumbnail models trained with different $\lambda_1$ and the bpp is 0.4.}
\label{fig:guide_loss}
\end{figure}
\noindent\textbf{Guidance Loss.}\quad
In Figure.\ref{fig:guide_loss}, we conduct a qualitative ablation study of guidance loss. 
It demonstrates that guidance loss is crucial to preserve the quality of LR images, without introducing noticeable degradation to HR images.

\noindent\textbf{Frequency-aware Decoder.}\quad
Because the efficiency of HR reconstruction is important for a better user experience, our decoder architecture has to be succinct and effective. 
In Table ~\ref{table:abla_arch}, we study the capacity of our decoder with frequency feature extractor $f$.
Removing $f$ in our framework(``Ours-w/o$f$") results in a drop in both the HR and LR RD performance.
Based on ``Ours-w/o$f$", we increase the residual blocks of the decoder from 16 to 22. 
``Ours-w/o$f$-b22" takes more upscaling time, but it ends up with a similar HR RD performance with ``Ours-w/ f" and a significantly inferior LR RD performance.
Since the spatial resolution of quantized coefficients $\hat{C}$ is $\frac{1}{8}$ of the embedding image $\hat{y}$, the running time of frequency feature extractor only accounts for $14.6\%$ of the entire decoder. Thus, our frequency feature extractor $f$ demonstrates a strong advantage with negligible computation cost. 

\noindent\textbf{Asymmetric encoder-decoder.}\quad
We quantitatively evaluate the influence of the encoder capacity on the RD performance in the bottom two rows of Tab.~\ref{table:abla_arch}. Based on ``Ours", We adjust the channel of our encoder from 64 to 48 and 96. The experiment shows that our framework benefits from the larger encoder. Since the 96-channel encoder is $2\times$ slower than 64 channel encoder and the improvement is marginal, we set encoder channel to 64 in most of our experiments to ease training.

Also, larger decoders can be applied to the same HyperThumbnail for better reconstruction quality. As shown in the table below, ``Ours-large'' decoder outperforms ``Ours-full'' decoder in the PSNR of HR significantly (Tab.~\ref{table:decoder-capacity}) with $4\times$ of parameters, sharing the same hyperthumbnails.

\begin{table}[h]
\centering
\resizebox{0.8\linewidth}{!}{
\begin{tabular}{@{}l@{\hspace{1mm}}c@{\hspace{1mm}}*{6}{c@{\hspace{1mm}}}r@{}}
\toprule
Decoder & Kodak & Set5 &  Set14  &  BSD100  &  Urb100 &  DIV2K\\
\midrule
Ours-full & 29.67 & 30.48 & 28.21 & 27.93 & 27.35 & 30.49\\
Ours-large & 29.74 & 30.56 & 28.39 & 28.01 & 27.74 & 30.61\\
\bottomrule
\end{tabular}
}
\caption{HR reconstruction PSNR with different decoder capacity.}
\label{table:decoder-capacity}
\vspace{-1.2em}   
\end{table}

\subsection{Additional qualitative results}
\begin{figure*}
    \centering
    \begin{tabular}{@{}c@{}c@{}c@{}c@{}c@{}c@{}c@{}}
    \footnotesize{EDSR~\cite{lim2017enhanced} ~$4\times$}&
    \footnotesize{SwinIR~\cite{liang2021swinir} ~$4\times$}&
    \footnotesize{IRN~\cite{xiao2020invertible} ~$4\times$}&
    \footnotesize{HCFlow~\cite{liang2021hierarchical}~$4\times$}&
    \footnotesize{Ours $4\times$}&
    \footnotesize{Ours-full $4\times$}& 
    \footnotesize{Ground Truth}
    \\
    \footnotesize{\& JPEG q=98}&
    \footnotesize{\& JPEG q=98}&
    \footnotesize{\& JPEG q=96}&
    \footnotesize{\& JPEG q=90}&
    \footnotesize{}&
    \footnotesize{}& 
    \footnotesize{}
    \\
    \resizebox{0.14285\linewidth}{!}{ }&
    \resizebox{0.14285\linewidth}{!}{ }&
    \resizebox{0.14285\linewidth}{!}{ }&
    \resizebox{0.14285\linewidth}{!}{ }&
    \resizebox{0.14285\linewidth}{!}{ }&
    \resizebox{0.14285\linewidth}{!}{ }&
    \resizebox{0.14285\linewidth}{!}{ }
    \vspace{-1em}
    \\
    \multicolumn{7}{@{}c@{}}{\includegraphics[width=1.0\linewidth]{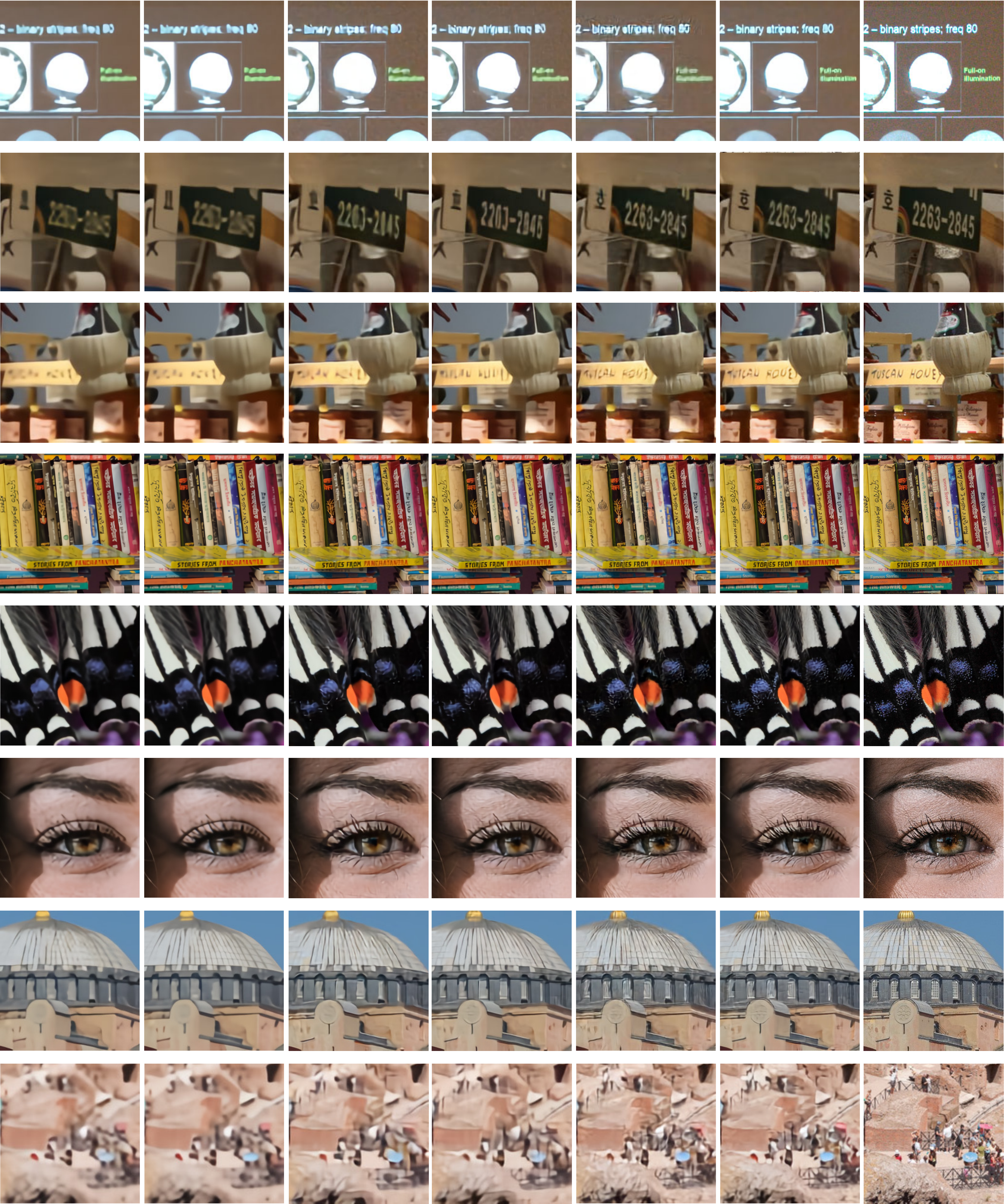}}
    \end{tabular}
    \caption{Visual results of performing $4\times$ rescaling on the DIV2K~\cite{agustsson2017ntire} and FiveK~\cite{bychkovsky2011learning} datasets with baseline methods and our models. The images are cropped to ease the comparison. Please zoom in for details.}
    \label{fig:supp_comparison}
\end{figure*}

\begin{figure*}
    \centering
    \vspace{-3em}
    \includegraphics[width=1.0\linewidth]
    {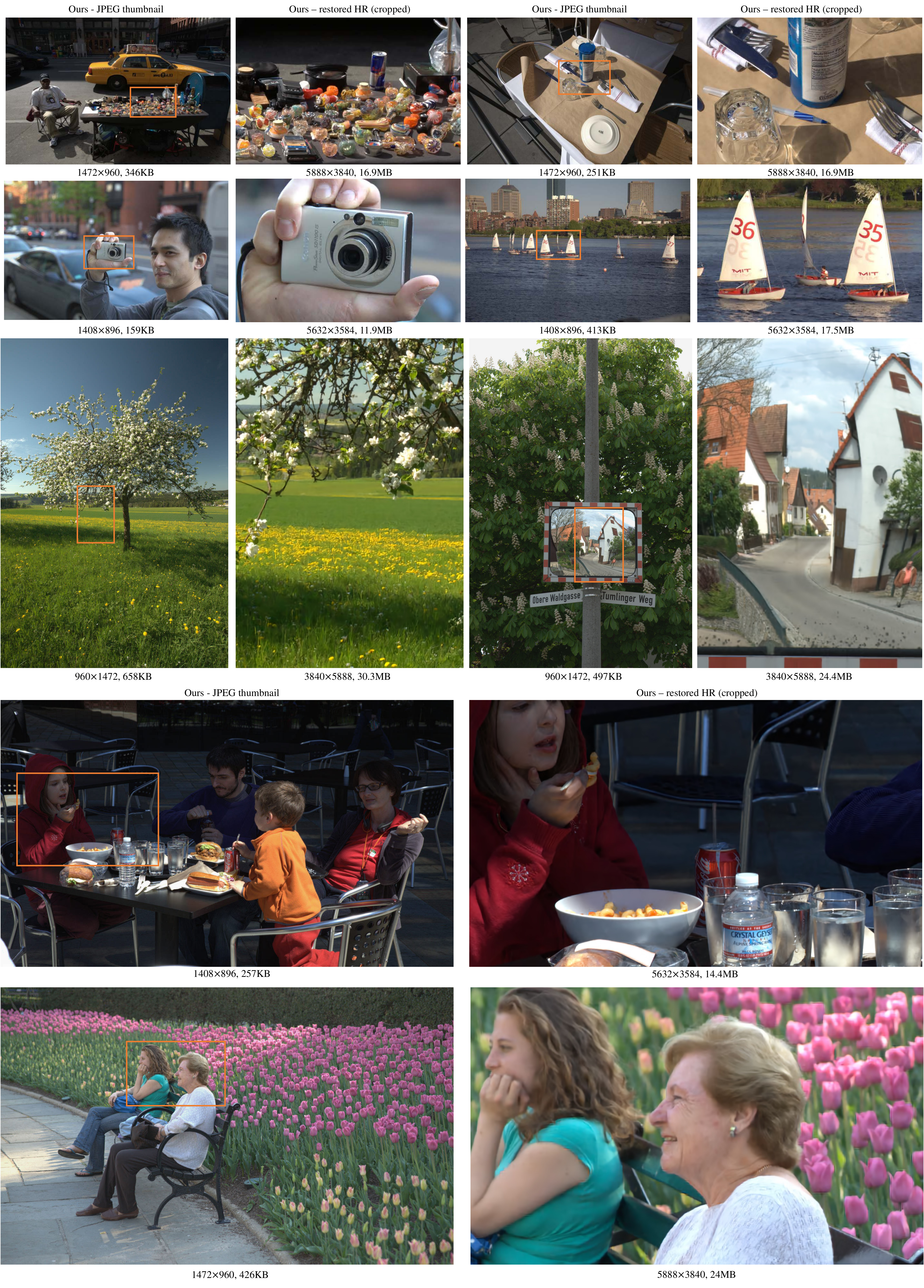}
    \caption{More results of $4\times$ rescaling with our framework on real-world 6K images~\cite{bychkovsky2011learning}. Please zoom in for details. Note that the images here are compressed due to the size limit of camera-ready.}
    \label{fig:6k_collection}
\end{figure*}

In this section, we visualize more results on the DIV2K~\cite{agustsson2017ntire} validation dataset and the FiveK~\cite{bychkovsky2011learning} dataset. Our model achieves the best balance between the embedding artifacts on LR and the restoration of HR detail. Our approach outperforms baseline methods, especially in texture restoration. All baseline models are trained on the same DIV2K training dataset to fit on guidance LR $\hat{y}$ and target HR $\hat{x}$. The results are cropped from the original image to ease comparison, please refer to Fig.~\ref{fig:supp_comparison}. Also, in Fig.~\ref{fig:6k_collection}, we visualize more rescaling results of real world 6K images with our framework.


\clearpage
%
%

\end{document}